\documentclass[11pt,a4paper]{article}
\usepackage[utf8]{inputenc}
\usepackage[T1]{fontenc}
\usepackage[margin=0.88in]{geometry}
\usepackage{lmodern}
\usepackage{microtype}
\usepackage{amsmath,amssymb,mathtools}
\usepackage{graphicx}
\usepackage{booktabs,multirow,array}
\usepackage[section]{placeins}
\usepackage[numbers,sort&compress]{natbib}
\usepackage{xcolor}
\usepackage{enumitem}
\usepackage{hyperref}

\hypersetup{
    colorlinks=true,
    linkcolor=blue!50!black,
    citecolor=teal!55!black,
    urlcolor=blue!50!black,
    pdftitle={Weaving 3D Physics and Property Models on a Self-Improving Molecular Loom},
    pdfauthor={Shangqian Chen, Guojiang Zhao, Yu Zeng, Xinye Li, Hongshuai Wang, Zhifeng Gao, Linfeng Zhang}
}

\title{\textbf{Weaving 3D Physics and Property Models on a Self-Improving Molecular Loom}}

\author{
Shangqian Chen, Guojiang Zhao, Yu Zeng, Xinye Li,\\
Hongshuai Wang, Zhifeng Gao, Linfeng Zhang\\[4pt]
\small DP Technology, Beijing, China\\
\footnotesize\texttt{\{chenshangqian, zhaogj, zengyu, lixinye, wanghongshuai, gaozf, zhanglf\}@dp.tech}
}

\date{November 2026}

\begin{document}
\maketitle

\begin{abstract}
Foundation models for three-dimensional molecular generation, property prediction, and universal machine-learning interatomic potentials have matured rapidly as standalone research artifacts. Translating these isolated models into reliable industrial discovery pipelines nevertheless remains difficult: organizations rely on heterogeneous black-box predictors that cannot be jointly retrained for each campaign, data-driven property surrogates suffer severe out-of-distribution errors that trigger reward hacking during reinforcement learning, and raw point-cloud generators exhibit cross-scale gradient interference alongside formal-charge artifacts. Here we introduce \textbf{Loom}, a composable framework that weaves reusable property predictors, hot-swappable interatomic potentials, and cross-paradigm 3D generative models into a physics-grounded recursive self-improvement loop without modifying external predictor weights. First, a unified asset protocol and declarative critic bus mount differentiable 3D networks, black-box QSAR ensembles, medicinal chemistry rules, and universal force fields behind monotonic applicability-domain and mechanical strain gates. Second, a heteroscedastic Gaussian-process atlas orthogonally decomposes generator dispersion from predictor epistemic uncertainty, restricting zero-cost inner reinforcement learning to trusted intervals while routing scaffold-diverse blindspot candidates to outer-loop quantum calibration via zero-retrain Bayesian residual wrappers. Third, scale-decoupled topological policy optimization and zero-displacement valence grounding ensure low-strain, closed-shell 3D structures across both octree autoregressive and flow-matching generators. Across scaffold-disjoint \texttt{OE62} benchmarks, \textbf{Loom} restores unit conditional calibration slope, eliminates spurious zwitterions without displacing heavy atoms, narrows cross-foundation surrogate hacking gaps by $4.6\times$ under synthetic accessibility constraints, and delivers $4.4\times$ to $9.65\times$ blind hit growth across offline and live \texttt{PBE0/def2-TZVP} self-improvement rounds. Under symmetric multi-tier \texttt{OE62} scaling, sparse octree weaving surpasses $21.6\times$ costlier flow-matching baselines in both blind discovery quality and unbiased \texttt{PBE0} quantum hit yield.
\end{abstract}

\section{Introduction}
\label{sec:intro}

Inverse design of functional three-dimensional (3D) organic molecules---spanning optoelectronic semiconductors \citep{stuke2020oe62,bhat2023ocelot,zhao2025virtual,zhao2026o2gen,wang2026mimflow}, catalysts, and drug-like scaffolds \citep{hoogeboom2022edm,xu2023geoldm,zeng2026propmolflow}---requires simultaneously satisfying closed-shell Lewis bonding, low-strain 3D equilibrium geometry, synthetic accessibility \citep{ertl2009sa}, and precise multi-property targets. While equivariant diffusion \citep{hoogeboom2022edm,xu2023geoldm}, flow-matching \citep{zeng2026propmolflow}, and octree autoregressive models \citep{lu2025uni3dar} have advanced unconditional 3D structure generation, bridging standalone academic models into an engineering discovery platform faces three systemic bottlenecks:

\begin{enumerate}[leftmargin=1.6em,itemsep=2pt,topsep=3pt]
    \item \textbf{Siloed Property and Physics Models Without Zero-Retrain Composition.} Practical pharmaceutical and materials teams maintain diverse, validated evaluators: coordinate-differentiable 3D networks such as \texttt{Uni-Mol} \citep{zhou2023unimol}, knowledge-enhanced and large black-box foundation predictors such as \texttt{Uni-Mol2} \citep{ji2024unimol2,zhao2025virtual}, proprietary 2D fingerprint QSAR ensembles, analytical chemistry filters \citep{ertl2009sa,bickerton2012qed}, and parametrized or universal machine-learning interatomic potentials (MLIPs) such as \texttt{DPA-4} \citep{zhao2025oscff,deepmodeling2026dpa4} and \texttt{HyFF} \citep{hyra2026hyff}. Existing conditional 3D pipelines typically hard-code specific benchmark targets into bespoke training scripts or require joint retraining across targets that lack co-labeled training tables.
    \item \textbf{Surrogate Blindspots and Unguided Self-Improvement.} Data-driven property predictors degrade sharply outside their historical training support, whereas universal interatomic potentials retain broad physical transferability across local bonding environments. On $4{,}737$ quantum-chemically evaluated generated conformers, frozen \texttt{Uni-Mol2 1.1B} exhibits $0.720\,\text{eV}$ \texttt{HOMO} mean absolute error (MAE) against \texttt{PBE0/def2-TZVP} density functional theory (DFT) due to strained geometries and charge artifacts, compared with $0.186\,\text{eV}$ for a DFT-calibrated \texttt{DPA-4} electronic head. When hit yield drops in a new target regime, naive reinforcement learning (RL) cannot distinguish \emph{generator under-exploration} from \emph{predictor epistemic blindness} or \emph{chemical prior ceilings}, either exploiting surrogate blindspots or wasting DFT and wet-lab budgets on strained, unsynthesizable structures.
    \item \textbf{Cross-Scale Gradient Interference and Point-Cloud Valence Fragility.} Delivering actionable candidates into an automated R\&D loop demands equilibrium 3D coordinates and clean closed-shell valence. In bondless 3D autoregressive generators, propagating molecule-level property rewards through fine $0.01\,\text{\AA}$ coordinate tokens injects high credit-assignment variance that degrades coarse topological steering. Meanwhile, omitting or misplacing a single peripheral hydrogen across voxel boundaries forces neutral bond perception such as \texttt{RDKit DetermineBonds} \citep{rdkit2026bonds} to assign spurious zwitterionic formal charges (\texttt{[C-]...[O+]}, \texttt{[B-]}, or halonium ions) on $15\%\text{--}39\%$ of otherwise valid scaffolds, corrupting both surrogate scoring and downstream SCF convergence.
\end{enumerate}

\begin{figure}[!t]
    \centering
    \includegraphics[width=\textwidth]{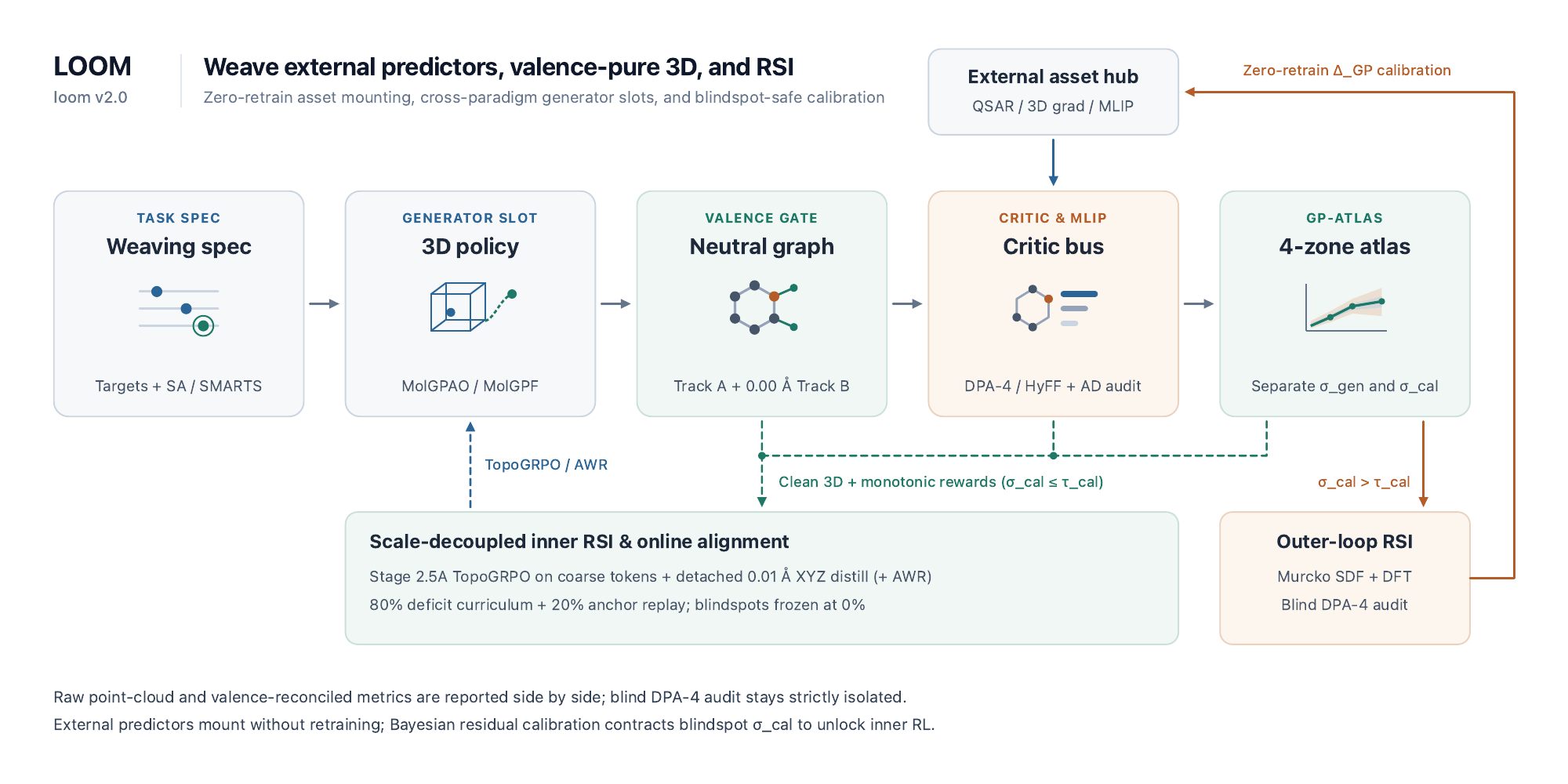}
    \caption{\textbf{Overview of the Loom architecture for weaving 3D physics and property models.} Declarative \texttt{TaskWeavingSpec} targets and inequality constraints drive interchangeable 3D \texttt{GeneratorSlot} backbones (\texttt{MolGPAO} \citep{chen2026molgpao} and flow-matching models \citep{zeng2026propmolflow,chen2026molgpf}), valence grounding, and a \texttt{CompositeCriticBus} backed without retraining by reusable property predictors, hot-swappable \texttt{DPA-4}/\texttt{HyFF} potentials, and applicability-domain auditors. \texttt{GPAtlas} separates generative dispersion $\sigma_{\text{gen}}$ from calibration uncertainty $\sigma_{\text{cal}}$: trusted regimes ($\sigma_{\text{cal}} \le \tau_{\text{cal}}$) execute zero-DFT inner alignment, whereas blindspots ($\sigma_{\text{cal}} > \tau_{\text{cal}}$) route Murcko-diverse scaffolds through MLIP relaxation to outer-loop DFT labeling and zero-retrain \texttt{ResidualCalibratedAsset} updates while keeping the blind \texttt{DPA-4} auditor strictly isolated.}
    \label{fig:architecture}
\end{figure}

To bridge this research-to-engineering gap, we present \textbf{Loom}, a composable 3D molecular weaving and physics-grounded recursive self-improvement (RSI) platform (Figure~\ref{fig:architecture}). Rather than training a monolithic predictor--generator supermodel from scratch, \textbf{Loom} pairs reusable property predictors (which define functional utility) with hot-swappable interatomic potentials \texttt{DPA-4} and \texttt{HyFF} (which enforce physical mechanics and stabilize geometries prior to quantum evaluation), weaving them across interchangeable 3D generative backbones \citep{lu2025uni3dar,chen2026molgpao,chen2026molgpf,zeng2026propmolflow}. Our main contributions are fourfold:

\begin{itemize}[leftmargin=1.5em,itemsep=2pt,topsep=3pt]
    \item \textbf{Composable Property and Physics Model Hub:} We design a unified \texttt{PredictorAsset} interface, hot-swappable \texttt{PotentialSlot}, and declarative \texttt{CompositeCriticBus} that mount differentiable 3D predictors, black-box QSAR models, zero-overhead inequality constraints, and universal MLIPs without retraining external assets, guarded by cross-oracle disagreement or single-oracle applicability-domain (AD) and atomic-force auditing.
    \item \textbf{Dual-Uncertainty Reachability Atlas (\texttt{GPAtlas}):} We construct 1D and 2D heteroscedastic Gaussian-process atlases \citep{rasmussen2006gp} that orthogonally separate generator control spread $\sigma_{\text{gen}}$ from calibration uncertainty $\sigma_{\text{cal}}$, mapping generative drift fields alongside narrow-gap surrogate failure zones and gating unreachable user queries prior to sampling.
    \item \textbf{Physics-Grounded Recursive Self-Improvement (RSI):} Using a four-quadrant diagnostic state machine, \textbf{Loom} restricts zero-cost inner-loop RL to trusted intervals, relaxes Murcko-diverse \citep{bemis1996murcko} blindspot candidates under MLIPs for outer-loop DFT acquisition, and updates a zero-retrain Bayesian residual wrapper (\texttt{ResidualCalibratedAsset}) whose contracted epistemic uncertainty causally unlocks former blindspots.
    \item \textbf{Cross-Paradigm Weaving and Physical-Valence Grounding:} Behind a unified \texttt{GeneratorSlot} and \texttt{LoomService} API spanning octree autoregression (\texttt{MolGPAO} \citep{chen2026molgpao}) and flow matching (\texttt{PropMolFlow} \citep{zeng2026propmolflow}, \texttt{MolGPF} \citep{chen2026molgpf}), we decouple coarse topological RL (\texttt{TopoGRPO}) from detached $0.01\,\text{\AA}$ equilibrium coordinate distillation and pair generation-side coordination guards with zero-heavy-displacement valence reconciliation.
\end{itemize}

\section{Problem Formulation and Surrogate Discrepancy}
\label{sec:prelim}

\subsection{Constrained Conditional 3D Molecular Design}
Let $\mathcal{M} = (\mathbf{Z}, \mathbf{R}, \mathcal{G})$ denote a 3D molecule comprising $N$ atoms with atomic numbers $\mathbf{Z} \in \mathbb{Z}_+^N$, Cartesian coordinates $\mathbf{R} \in \mathbb{R}^{N \times 3}$, and an inferred closed-shell Lewis graph $\mathcal{G} = (\mathcal{V}, \mathcal{E}, \mathbf{b}, \mathbf{q})$ with integer bond orders $\mathbf{b}$ and formal charges $\mathbf{q} \in \mathbb{Z}^N$ satisfying $\sum_{i=1}^N q_i = 0$. Given target specifications $\mathbf{t} = (t_1, \dots, t_K) \in \mathbb{R}^K$ with missingness mask $\mathbf{m} \in \{0,1\}^K$, tolerances $\boldsymbol{\tau} = (\tau_1, \dots, \tau_K)$, and $M$ auxiliary inequality constraints $\{g_j(\mathcal{M}) \le c_j\}_{j=1}^M$ such as synthetic accessibility $\text{SA}(\mathcal{M}) \le 3.5$ or reactive-group count $\text{SMARTS}(\mathcal{M}) \le 0$, we seek a conditional policy $\pi_\theta(\mathcal{M} \mid \mathbf{t}, \mathbf{m})$ maximizing the calibrated, constraint-satisfying, valence-clean hit yield:
\begin{equation}
\label{eq:problem}
\begin{aligned}
\mathcal{Y}_{\text{clean}}(\mathbf{t}) = \mathbb{E}_{\mathcal{M} \sim \pi_\theta(\cdot \mid \mathbf{t}, \mathbf{m})} \!\Bigg[ \mathbb{I}\!\Bigg(
&\mathcal{M} \in \mathcal{V}_{\text{3D}} \;\land\; \mathcal{Q}_{\text{exotic}}(\mathcal{M}) = \emptyset \\
&\land\; \bigcap_{j=1}^M \{g_j(\mathcal{M}) \le c_j\} \;\land\; \bigcap_{k: m_k=1} \big\{|y_k^\star(\mathcal{M}) - t_k| \le \tau_k\big\}
\Bigg) \Bigg],
\end{aligned}
\end{equation}
where $\mathcal{V}_{\text{3D}}$ denotes single-component 3D covalent validity surviving MLIP relaxation without bond rearrangement, $\mathcal{Q}_{\text{exotic}}(\mathcal{M})$ denotes unauthorized formal charges outside benign dipolar groups, and $y_k^\star(\mathcal{M})$ is the true quantum-chemical or experimental observable.

\subsection{Quantum Reference Benchmark and Surrogate Discrepancy}
\label{sec:quantum_audit}
A central premise of \textbf{Loom} is that a single surrogate oracle used to optimize $\pi_\theta$ cannot simultaneously serve as its own unbiased evaluator. To quantify surrogate vulnerability on generated 3D geometries, we consolidated historical \texttt{PBE0/def2-TZVP/DF} single-point DFT calculations \citep{sun2020pyscf,li2024gpu4pyscf,stuke2020oe62} across generated and reference cohorts while excluding all canonical molecular graphs and Murcko scaffolds overlapping validation or test splits. This yielded $4{,}737$ leak-free generated DFT structures ($3{,}949$ train, $398$ validation, and $390$ test), which we merged with $43{,}858$ scaffold-disjoint \texttt{OE62} training structures into a $47{,}807$-conformer pool spanning $12$ elements up to $80$ atoms.

Evaluating two distinct 3D model families on the $390$ held-out generated DFT test conformers exposes a sharp physical asymmetry. Frozen \texttt{Uni-Mol2 1.1B} \citep{ji2024unimol2} records test \texttt{HOMO}/\texttt{LUMO} MAEs of $0.720 / 0.778\,\text{eV}$ and root-mean-square errors (RMSEs) of $1.106 / 1.253\,\text{eV}$ due to sensitivity to bond-length strain and formal-charge artifacts. By contrast, an independent DFT-calibrated \texttt{DPA-4} electronic head \citep{deepmodeling2026dpa4} achieves $0.186 / 0.183\,\text{eV}$ MAE and $0.268 / 0.268\,\text{eV}$ RMSE. Throughout all experiments, \texttt{Uni-Mol2} or a restricted historical QSAR ensemble serves exclusively on the training side as the reward surrogate, whereas the calibrated \texttt{DPA-4} electronic head remains strictly isolated as a blind \texttt{AuditOracle} that never participates in relabeling, filtering, or policy optimization.

\section{The Molecular Loom Architecture}
\label{sec:method}

\subsection{Composable Property and Physics Model Hub}
\label{sec:asset_bus}

\paragraph{Reusable Property Assets and Hot-Swappable Physics Potentials.}
To reuse existing enterprise predictors alongside universal physics models without task-specific hard-coding, \textbf{Loom} mounts evaluators under a uniform \texttt{PredictorAsset} interface declaring input modality, coordinate differentiability, native uncertainty support, and physical units, paired with a dedicated \texttt{PotentialSlot} for interatomic mechanics:
\begin{enumerate}[leftmargin=1.6em,itemsep=1pt,topsep=2pt]
    \item \textbf{Differentiable 3D Property Models (\texttt{UniMol3DAsset}):} Wraps coordinate-differentiable 3D networks \citep{zhou2023unimol} to expose both batched scalar predictions $\hat{\mathbf{y}}(\mathcal{M})$ and analytical Cartesian gradients $\nabla_{\mathbf{R}}\hat{y}_k(\mathbf{Z}, \mathbf{R})$ via autograd on pairwise distances;
    \item \textbf{Black-Box 3D and 2D QSAR Ensembles (\texttt{SlotOracleAsset}, \texttt{FingerprintQSARAsset}, \texttt{CallablePredictorAsset}):} Wraps frozen 3D foundation models (\texttt{Uni-Mol2}), $K$-fold Morgan fingerprint ensembles returning ensemble standard deviation $\sigma_{\text{ens}}$, or external \texttt{ONNX} and \texttt{scikit-learn} predictors;
    \item \textbf{Analytical Chemistry Rules (\texttt{BuiltInRuleAsset}):} Evaluates deterministic descriptors including \texttt{SA\_Score} \citep{ertl2009sa}, \texttt{QED} \citep{bickerton2012qed}, molecular weight, rotatable bonds, maximum ring size, strained $3/4$-membered ring counts, and reactive \texttt{SMARTS} alerts;
    \item \textbf{Hot-Swappable Interatomic Potentials (\texttt{PotentialSlot}):} Unifies universal MLIPs---\texttt{DPA-4} \citep{deepmodeling2026dpa4} and \texttt{HyFF vivace} \citep{hyra2026hyff}---behind spatially separated \texttt{float64} batch packing, which reduces batch-$32$ force numerical drift from $7.09 \times 10^{-3}$ to $7.89 \times 10^{-6}\,\text{eV/\AA}$. \texttt{PotentialSlot} supplies zero-step maximum atomic force $F_{\max} = \max_a \|\mathbf{F}_a\|_2$, batched \texttt{FIRE} geometry relaxation, covalent graph-preservation checks, and Kabsch relaxation deformation $\text{RMSD}_{\text{Kabsch}}$, acting simultaneously as an anti-hacking mechanical filter, an equilibrium distillation teacher, and a pre-DFT stabilizer.
\end{enumerate}

\paragraph{Physics-Constrained Coordinate Refinement.}
When a mounted property model exposes coordinate gradients, \texttt{PhysicsConstrainedCoordRefiner} jointly optimizes 3D conformers on the composite objective:
\begin{equation}
\label{eq:coord_refine}
\min_{\mathbf{R} \in \mathbb{R}^{N \times 3}} \; E_{\text{MLIP}}(\mathbf{Z}, \mathbf{R}) + \lambda \sum_{k=1}^K w_k \, \mathcal{L}_{\text{obj}, k}\!\big(\hat{y}_k(\mathbf{Z}, \mathbf{R}); t_k, \tau_k\big),
\end{equation}
where interatomic forces $-\nabla_{\mathbf{R}} E_{\text{MLIP}}$ preserve equilibrium bond lengths and angles while $-\lambda \nabla_{\mathbf{R}}\mathcal{L}_{\text{obj}, k}$ steers soft conjugated torsional degrees of freedom \citep{zhao2025oscff} toward target windows under strict graph-preservation and $F_{\max}$ acceptance gates.

\paragraph{Declarative Task Weaving and Monotonic Critic Bus.}
\texttt{CompositeCriticBus} compiles declarative YAML specifications (\texttt{TaskWeavingSpec}) into three orthogonal channels: (i) \textbf{Objectives} (\texttt{point}, \texttt{interval}, \texttt{max}, or \texttt{min}), which configure the generator conditioning dimension $K$ and impute missing multi-model pseudo-labels across unlabeled conformer pools; (ii) \textbf{Inequality Constraints} ($g_j(\mathcal{M}) \le c_j$), which gate synthetic and structural feasibility without consuming conditioning dimensions; and (iii) \textbf{Anti-Hacking Auditors}. When an independent second electronic oracle is unavailable---the norm in proprietary QSAR workflows---the bus activates \texttt{single\_oracle\_ad} mode, combining nearest-neighbor Morgan Tanimoto applicability-domain distance $d_{\text{AD}}(\mathcal{M}) = 1 - \max_{\mathcal{M}' \in \mathcal{D}_{\text{ref}}} \text{Sim}(\mathcal{M}, \mathcal{M}')$, ensemble variance $\sigma_{\text{ens}}(\mathcal{M})$, and \texttt{PotentialSlot} atomic-force gates. Across online rollouts, \texttt{CompositeCriticBus} enforces a strict monotonic reward hierarchy:
\begin{equation}
\label{eq:hierarchy}
R_{\text{clean+constr+hit}} > R_{\text{clean+constr+miss}} > R_{\text{soft-constr-viol}} > R_{\text{exotic-charge / high-strain}} > R_{\text{illegal-3D}},
\end{equation}
guaranteeing that surrogate outliers never outrank valence-clean, low-strain, constraint-satisfying molecules.

\subsection{Uncertainty Decomposition and Recursive Self-Improvement}
\label{sec:rsi}

\paragraph{Zero-Retrain Bayesian Residual Calibration.}
When an external black-box predictor $\hat{y}_{\text{base}}$ errs outside its historical domain or lacks calibrated epistemic uncertainty, \texttt{Residual\allowbreak Calibrated\allowbreak Asset} wraps it without altering any base parameters. Each molecule is mapped to a $29$-dimensional explicit-H-invariant descriptor $\boldsymbol{\phi}(\mathcal{M}) \in \mathbb{R}^{29}$ combining $16$ folded Morgan bit-density channels, heavy-atom stoichiometry, $sp^3$ and aromatic heavy-atom fractions, $N_{\text{H}}/N_{\text{heavy}}$ saturation, $N_{\text{total}}/50$ molecular scale, and heavy-atom 3D radius of gyration and inertial anisotropy. Over external reference labels $\{(\mathcal{M}_i, y_i^{\text{true}})\}$, \texttt{Incremental\allowbreak Calibrator} fits a global linear Ridge drift $f_{\text{ridge}}(\boldsymbol{\phi})$ plus a fixed-prior-amplitude $\text{Mat\'ern}(\nu=3/2) + \text{WhiteKernel}$ Gaussian process $\delta_{\text{GP}}(\boldsymbol{\phi})$ \citep{rasmussen2006gp} on detrended residuals:
\begin{equation}
\label{eq:rescal}
\begin{aligned}
\hat{y}_{\text{cal}}(\mathcal{M}) &= \hat{y}_{\text{base}}(\mathcal{M}) + f_{\text{ridge}}\!\big(\boldsymbol{\phi}(\mathcal{M})\big) + \delta_{\text{GP}}\!\big(\boldsymbol{\phi}(\mathcal{M})\big), \\
\sigma_{\text{cal}}(\mathcal{M}) &= \sqrt{\max\!\Big(\sigma_{\text{GP}}^2\big(\boldsymbol{\phi}(\mathcal{M})\big) - \sigma_{\text{noise}}^2,\, 0\Big)}.
\end{aligned}
\end{equation}
Subtracting observation noise $\sigma_{\text{noise}}^2$ isolates epistemic calibration uncertainty $\sigma_{\text{cal}}(\mathcal{M})$, which contracts monotonically below a trust cap $\tau_{\text{cal}}$ in chemical sub-regions populated by new DFT or experimental measurements.

\paragraph{Dual-Uncertainty Reachability Atlas (\texttt{GPAtlas}).}
Over a target grid $\mathcal{T} \subset \mathbb{R}^D$ ($D \in \{1, 2\}$), \texttt{HeteroscedasticGPAtlas} fits Gaussian-process surfaces for achieved conditional mean $\hat{\mu}_{\text{gen}}(\mathbf{t})$, intrinsic generator dispersion $\sigma_{\text{gen}}(\mathbf{t})$, calibration uncertainty $\sigma_{\text{cal}}(\mathbf{t})$, and evaluability rate $p_{\text{eval}}(\mathbf{t})$. Integrating over orthogonal total variance $\sigma_{\text{total}}^2(\mathbf{t}) = \sigma_{\text{gen}}^2(\mathbf{t}) + \sigma_{\text{cal}}^2(\mathbf{t})$ yields the calibrated hit probability within $[t_k - \tau_k, t_k + \tau_k]$:
\begin{equation}
\label{eq:phit}
P_{\text{hit}}^{\text{cal}}(\mathbf{t}) = p_{\text{eval}}(\mathbf{t}) \prod_{k=1}^D \left[ \Phi\!\left(\frac{t_k + \tau_k - \hat{\mu}_{\text{gen}, k}(\mathbf{t})}{\sqrt{\sigma_{\text{gen}, k}^2(\mathbf{t}) + \sigma_{\text{cal}, k}^2(\mathbf{t})}}\right) - \Phi\!\left(\frac{t_k - \tau_k - \hat{\mu}_{\text{gen}, k}(\mathbf{t})}{\sqrt{\sigma_{\text{gen}, k}^2(\mathbf{t}) + \sigma_{\text{cal}, k}^2(\mathbf{t})}}\right) \right].
\end{equation}
Before generation begins, a pre-generation gatekeeper (\texttt{recommend\_target}) flags untrusted or unreachable queries and suggests the nearest viable target.

\paragraph{Four-Quadrant Diagnostic Flywheel.}
At each RSI cycle, \texttt{RSIDiagnoser} partitions target cells $\mathbf{t} \in \mathcal{T}$ into four causal states:
\begin{enumerate}[leftmargin=1.6em,itemsep=2pt,topsep=2pt]
    \item \textbf{\texttt{PRIOR\_CEILING}:} Vanishing source prior density $\rho_{\text{src}}(\mathbf{t}) < \rho_{\min}$ paired with strong generative pull $|\hat{\mu}_{\text{gen}}(\mathbf{t}) - \mathbf{t}| > \tau_{\text{ceil}}$ assigns $0\%$ RL weight, avoiding futile optimization beyond stable neutral organic space;
    \item \textbf{\texttt{OUTER\_BLINDSPOT}:} Elevated epistemic uncertainty $\sigma_{\text{cal}}(\mathbf{t}) > \tau_{\text{cal}}$ freezes inner-loop RL weight at $0\%$ to prevent reward hacking and triggers \texttt{OuterRSIAcquisition}, selecting clean, constraint-satisfying candidates by Murcko-diversified \citep{bemis1996murcko} acquisition utility:
    \begin{equation}
    \label{eq:acq}
    u(\mathcal{M}) = \sigma_{\text{cal}}(\mathcal{M}) \cdot \mathbb{I}\big[\text{Clean}(\mathcal{M}) \land \text{ConstraintsMet}(\mathcal{M})\big] \cdot d_{\text{Murcko}}(\mathcal{M}, \mathcal{S}_{\text{selected}}),
    \end{equation}
    relaxing selected structures via \texttt{PotentialSlot} for outer-loop DFT or wet-lab evaluation;
    \item \textbf{\texttt{INNER\_UNDEREXPLORED}:} Trusted calibration ($\sigma_{\text{cal}}(\mathbf{t}) \le \tau_{\text{cal}}$) with unsaturated hit yield ($P_{\text{hit}}^{\text{cal}}(\mathbf{t}) < p_{\text{conv}}$) receives $80\%$ of curriculum RL budget weighted by hit deficit $1 - P_{\text{hit}}^{\text{cal}}(\mathbf{t})$, running zero-DFT \texttt{InnerRSIRunner} updates guarded by a non-regression acceptance check;
    \item \textbf{\texttt{CONVERGED}:} Trusted cells meeting $P_{\text{hit}}^{\text{cal}}(\mathbf{t}) \ge p_{\text{conv}}$ retain a $20\%$ uniform anchor replay budget across trusted quantiles to prevent catastrophic forgetting.
\end{enumerate}

\subsection{Cross-Paradigm Generation and Physical Grounding}
\label{sec:scale_decoupled}
\label{sec:valence}

\paragraph{Generator-Slot Abstraction and Self-Driving Service API.}
\textbf{Loom} isolates generative paradigm mechanics behind \texttt{GeneratorSlot}. \texttt{LoomUni3DARGenerator} mounts octree autoregressive backbones---\texttt{Uni-3DAR} \citep{lu2025uni3dar} and the speculative octree engine \texttt{MolGPAO} \citep{chen2026molgpao}, which incorporates two-stage process-critic pruning, sub-millisecond $SE(3)$ Morton-seam healing (\texttt{PostTreeGraphRefiner}), scale-free atom-budget modulation without $N_{\text{atoms}}$ property leakage, and stride-$2$ GPU leaf decoding. \texttt{LoomFlowMolGenerator} mounts continuous--discrete $SE(3)$ flow matching (\texttt{PropMolFlow} \citep{zeng2026propmolflow} and \texttt{MolGPF} \citep{chen2026molgpf}) with lightweight online advantage-weighted regression (\texttt{AWR}). Above \texttt{GeneratorSlot}, \texttt{LoomService} exposes a declarative three-pillar self-driving laboratory API: standardized effort tiers ($M \in \{1, 2, 4, 8, 12, 16\}$) ranked by closed-form Bayesian goal-satisfaction probability, asynchronous $K$-dimensional partial-\texttt{NaN} assay ingestion (\texttt{ingest\_feedback}) with multi-dimensional diagnosis (\texttt{diagnose\_nd}), and capability-adaptive policy alignment (\texttt{TopoGRPO} for exact-likelihood models and online \texttt{AWR} for flow models).

\paragraph{Unbounded Conditioning and Scale-Decoupled Alignment.}
In early \texttt{Uni-3DAR} adapters, per-layer adaptive LayerNorm gates clipped by $0.08 \tanh(\cdot)$ with frozen attention layers capped conditional calibration slopes at $0.41\text{--}0.54$. Removing the $\tanh$ bottleneck, injecting continuous Fourier features of $(\mathbf{t}, \mathbf{m})$ with rank-$16$ \texttt{LoRA} \citep{hu2022lora} across all $24$ Transformer blocks, training with conditional dropout ($p_{\text{drop}} = 0.15$) plus contrastive conditioning loss, and applying tree-depth-annealed classifier-free guidance (\texttt{CFG} \citep{ho2022cfg}) restores full dynamic range:
\begin{equation}
\label{eq:cfg}
\ell_{\text{CFG}}(u \mid d) = \ell_{\text{uncond}}(u) + \Big(1 + (s_{\text{CFG}} - 1)\,\gamma^{\max(d - d_0, 0)}\Big)\big(\ell_{\text{cond}}(u) - \ell_{\text{uncond}}(u)\big),
\end{equation}
steering coarse scaffold octree levels ($d \le d_0$) strongly while leaving sub-voxel leaf offsets unguided to preserve equilibrium bond lengths.

Crucially, octree leaf atoms emit one coarse element token ($Z_i$) alongside three $0.01\,\text{\AA}$ coordinate tokens ($\Delta x_i, \Delta y_i, \Delta z_i$). Propagating scalar property rewards across all tokens injects $3\times$ spatial token credit-assignment variance into shared representations, degrading within-target precision. \textbf{Loom} therefore decouples policy alignment across spatial scales:
\begin{enumerate}[leftmargin=1.6em,itemsep=2pt,topsep=2pt]
    \item \textbf{Stage 2.5A (\texttt{TopoGRPO}):} Group-relative policy optimization \citep{shao2024grpo} ($G = 8\text{--}16$) restricts policy gradients strictly to coarse octree-split and element tokens ($\mathcal{T}_{\text{topo}}$), zeroing coordinate token gradients and applying dual-finite token-level KL divergence to prevent hard-mask \texttt{NaN} instabilities:
    \begin{equation}
    \label{eq:topogrpo}
    \begin{aligned}
    \mathcal{L}_{\text{TopoGRPO}}(\theta) = -\mathbb{E}_{g, i}\!\Bigg[\frac{1}{|\mathcal{T}_{\text{topo}}^{(i)}|} \sum_{u \in \mathcal{T}_{\text{topo}}^{(i)}} \Big( &\min\!\big(\rho_u A_i, \text{clip}(\rho_u, 1-\epsilon, 1+\epsilon)A_i\big) \\
    &-\; \beta_{\text{KL}} D_{\text{KL}}^{\text{finite}}(\pi_\theta \| \pi_{\text{ref}})_u \Big)\Bigg] + \lambda_{\text{rep}} \mathcal{L}_{\text{SFT}}^{\text{replay}}.
    \end{aligned}
    \end{equation}
    \item \textbf{Stage 2.5B (Detached Coordinate Distillation):} Detaching Transformer hidden states, lightweight rank-$8$ \texttt{LoRA} adapters ($111{,}744$ parameters) on the three $0.01\,\text{\AA}$ \texttt{XYZ} heads are supervised on $40$-step MLIP-relaxed equilibrium conformers, lowering zero-step strain without perturbing topological calibration.
\end{enumerate}

\paragraph{Dual-Track Valence Grounding for Closed-Loop Delivery.}
Unaligned 3D point clouds frequently acquire spurious zwitterions (\texttt{[C-]...[O+]}, \texttt{[B-]}, halonium) when a carbonyl oxygen receives a third close neighbor ($< 1.65\,\text{\AA}$) or a peripheral hydrogen shifts across voxel boundaries during \texttt{DetermineBonds(..., charge=0)} \citep{rdkit2026bonds}. To ensure every candidate entering surrogate scoring or DFT acquisition is chemically grounded, we combine two tracks and report both \emph{Raw Point-Cloud} and \emph{Valence-Reconciled} metrics side by side:
\begin{itemize}[leftmargin=1.5em,itemsep=2pt,topsep=2pt]
    \item \textbf{Track A (Generation-Side Coordination Guard):} During batched GPU decoding, pre-softmax element logits mask additional bonding neighbors once oxygen has a short double bond ($d < 1.28\,\text{\AA}$) or two covalent neighbors, halogens reach coordination $1$, or boron reaches neutral coordination $3$. Simultaneously, a lightweight linear \texttt{HeavyAtomValenceHead} ($< 30\text{K}$ parameters) supervises per-heavy-atom hydrogen counts $n_{\text{H}} \in \{0,\dots,4\}$ and formal charges $q \in \{-1, 0, +1\}$.
    \item \textbf{Track B (Zero-Displacement \texttt{ValenceReconciler}):} Valid raw molecules with neutral valence or whitelisted dipolar groups (nitro, $N$-oxide, sulfoxide, diazo/azide) pass untouched. Only when unauthorized formal charges or peripheral-H mismatches arise on an intact heavy-atom skeleton, \texttt{ValenceReconciler} locks all heavy-atom coordinates identically ($\text{RMSD}_{\text{heavy}} \equiv 0.0000\,\text{\AA}$), assigns neutral closed-shell Lewis bond orders from 3D geometry, rebuilds explicit hydrogens, and runs $10$ batched H-only MLIP relaxation steps.
\end{itemize}

\section{Experimental Validation}
\label{sec:experiments}

\subsection{Experimental Setup and Inference Efficiency}
\label{sec:exp_setup}

Primary benchmarks are conducted on the scaffold-disjoint \texttt{OE62} organic semiconductor dataset \citep{stuke2020oe62} merged with our leak-free consolidated DFT pool ($47{,}807$ training conformers across $\text{H, B, C, N, O, F, Si, P, S, Cl, Br, I}$). Every end-to-end evaluation uses fixed random streams across $7$ target quantiles ($P_{10}, P_{25}, P_{35}, P_{50}, P_{65}, P_{75}, P_{90}$) $\times 256$ samples $= 1{,}792$ uncurated proposals with hit tolerance $\tau = \pm 0.25\,\text{eV}$; non-evaluable or invalid structures are never excluded from hit-rate denominators.

Table~\ref{tab:speed} benchmarks single-GPU throughput on an \texttt{NVIDIA A800-80GB}. Vectorized \texttt{BF16} octree decoding with fused scaled dot-product attention reaches $32.52\,\text{mol/s}$ at batch $256$ ($7.29\times$ faster than the legacy per-row \texttt{FP32} loop) within $3.23\,\text{GB}$ peak VRAM, and sustains $22.25\text{--}29.51\,\text{mol/s}$ with dual-branch \texttt{CFG} and GPU coordination guards enabled. Training throughput reaches $80.92\,\text{mol/s}$ with $2\times$-pass contrastive conditioning.

\begin{table}[!t]
\centering
\caption{Single-GPU sampling and training throughput across precision, batch size, and guidance settings on an \texttt{NVIDIA A800-80GB}.}
\label{tab:speed}
\resizebox{\textwidth}{!}{%
\begin{tabular}{llrrrr}
\toprule
Pipeline Configuration & Precision & Batch & CFG Scale & Peak VRAM (GB) & Throughput (mol/s) \\
\midrule
Legacy baseline pipeline (per-row \texttt{FP32} loop) & \texttt{FP32} & 32 & 1.00 & --- & 4.46 \\
Vectorized \texttt{FP32} row-streams baseline & \texttt{FP32} & 32 & 1.00 & --- & 6.41 \\
\textbf{Loom} (\texttt{BF16} fused octree decoding) & \texttt{BF16} & 64 & 1.00 & 1.92 & 11.75 \\
\textbf{Loom} (\texttt{BF16} fused octree decoding) & \texttt{BF16} & 128 & 1.00 & 2.39 & 20.19 \\
\textbf{Loom} (\texttt{BF16} fused octree decoding) & \texttt{BF16} & 256 & 1.00 & 3.23 & \textbf{32.52} ($7.29\times$) \\
\textbf{Loom} (dual-branch \texttt{CFG} + geometric guard) & \texttt{BF16} & 128--256 & 1.75--2.00 & 3.36 & \textbf{22.25--29.51} \\
\midrule
Training: Legacy \texttt{FP32} SFT (single pass, no contrast) & \texttt{FP32} & 32 & --- & --- & 72.85 \\
Training: \textbf{Loom} \texttt{BF16} \texttt{LoRA-16} + $2\times$ contrastive pass & \texttt{BF16} & 32 & --- & --- & \textbf{80.92} ($1.11\times$) \\
\bottomrule
\end{tabular}%
}
\vspace{2pt}\par{\footnotesize Sampling throughput is measured over $256$ generated molecules per arm; training throughput measures end-to-end optimizer steps.}
\end{table}

\subsection{Alignment Progression and Physics-Model Hot-Swapping}
\label{sec:exp_progression}

\begin{figure}[!t]
    \centering
    \includegraphics[width=\textwidth]{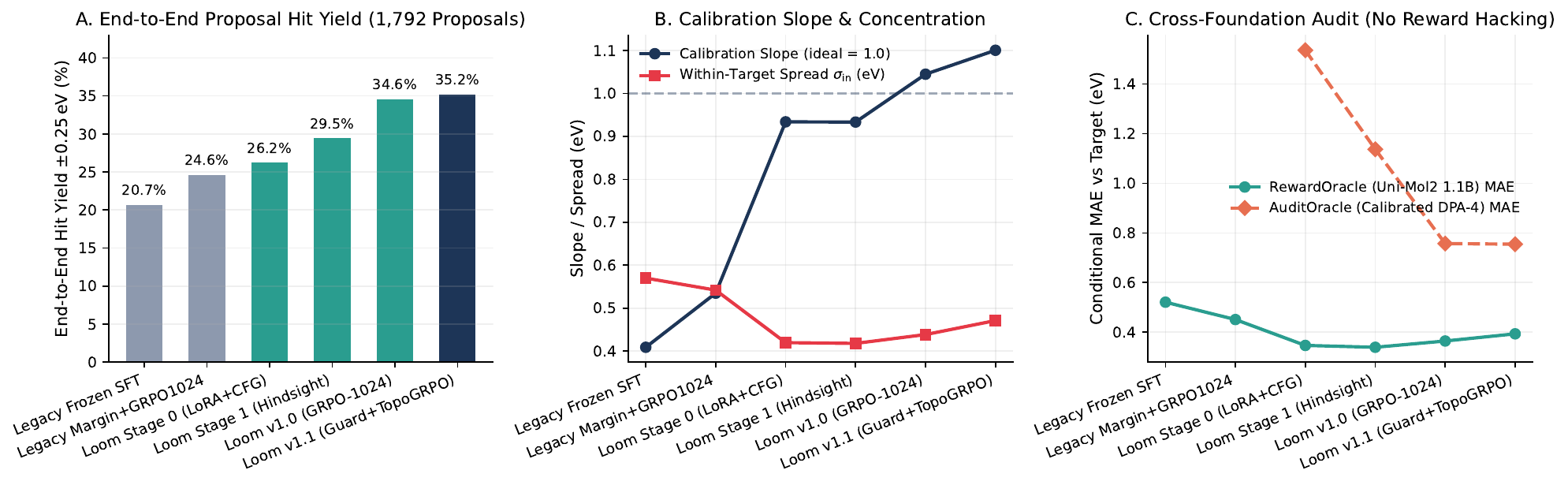}
    \caption{\textbf{Stagewise alignment progression on \texttt{OE62 HOMO} ($1{,}792$ proposals).} (\textbf{A}) Total and zwitterion-clean hit yields within $\pm 0.25\,\text{eV}$. (\textbf{B}) Conditional calibration slope and within-target dispersion $\sigma_{\text{in}}$ after removing $\tanh$ conditioning compression. (\textbf{C}) Isolated cross-foundation \texttt{AuditOracle} MAE across alignment stages.}
    \label{fig:progression}
\end{figure}

Table~\ref{tab:progression} and Figure~\ref{fig:progression} trace end-to-end \texttt{OE62 HOMO} generation from legacy frozen conditioners through Stage~2.5A+2.5B. Removing the $0.08\tanh$ conditioning gate in Stage~0 immediately lifts conditional calibration slope from $0.409$ to $0.934$ and tightens within-target spread $\sigma_{\text{in}}$ from $0.570$ to $0.420\,\text{eV}$, although unaligned zero-step legality starts at $59.6\%$ with $1.536\,\text{eV}$ blind \texttt{AuditOracle} MAE. Adding \texttt{DPA-4} graph-preserving Hindsight relabeling (Stage~1) and monotonic \texttt{GRPO} (Stage~2) raises hit yield to $34.60\%$ ($620/1{,}792$) while cutting post-relaxation blind \texttt{Audit} MAE to $0.757\,\text{eV}$. Combining generation-side coordination guards with \texttt{TopoGRPO} (Stage~2.5A) reaches $35.21\%$ hit yield ($631/1{,}792$, comprising $621$ novel unique graphs), $451$ zwitterion-clean hits, $84.6\%$ raw legality ($85.4\%$ with Stage~2.5B), and $0.744\,\text{eV}$ blind \texttt{Audit} MAE.

\begin{table}[!t]
\centering
\caption{Stagewise conditional \texttt{OE62 HOMO} generation across $1{,}792$ unedited raw point-cloud proposals ($\pm 0.25\,\text{eV}$ hit window).}
\label{tab:progression}
\resizebox{\textwidth}{!}{%
\begin{tabular}{lrrrrrrrr}
\toprule
Method / Stage & Hits / 1792 & Hit Yield & Clean Hits & 3D Legal & Slope & $\sigma_{\text{in}}$ (eV) & Reward MAE & Blind Audit MAE \\
\midrule
Legacy Frozen SFT & 371 & 20.70\% & --- & 72.1\% & 0.409 & 0.570 & 0.521 & --- \\
Legacy Margin + \texttt{GRPO-1024} & 441 & 24.61\% & --- & 70.0\% & 0.535 & 0.542 & 0.451 & --- \\
\textbf{Loom} Stage 0 (\texttt{LoRA-16} + \texttt{CFG=1.75}) & 470 & 26.23\% & 184 (10.3\%) & 59.6\% & 0.934 & 0.420 & 0.347 & 1.536 \\
\textbf{Loom} Stage 1 (\texttt{DPA-4} Hindsight) & 528 & 29.46\% & 286 (16.0\%) & 63.6\% & 0.933 & \textbf{0.418} & \textbf{0.339} & 1.137 \\
\textbf{Loom v1.0} Stage 2 (\texttt{GRPO-1024}, \texttt{CFG=2.0}) & 620 & 34.60\% & 449 (25.1\%) & 80.5\% & 1.045 & 0.439 & 0.364 & 0.757 \\
\textbf{Loom v1.1} \texttt{GRPO-1024} + Guard + \texttt{XYZ} Distill & 625 & 34.88\% & 450 (25.1\%) & 80.1\% & 1.037 & 0.427 & 0.353 & 0.825 \\
\textbf{Loom v1.1} Stage 2.5A (\texttt{Guard+TopoGRPO}, \texttt{CFG=1.75}) & 604 & 33.71\% & 436 (24.3\%) & 84.6\% & \textbf{0.986} & 0.480 & 0.403 & \textbf{0.744} \\
\textbf{Loom v1.1} Stage 2.5A (\texttt{Guard+TopoGRPO}, \texttt{CFG=2.0}) & \textbf{631} & \textbf{35.21\%} & \textbf{451 (25.2\%)} & 83.5\% & 1.100 & 0.471 & 0.393 & 0.755 \\
\textbf{Loom v1.1} Stage 2.5A+2.5B (\texttt{TopoGRPO+XYZ}, \texttt{CFG=1.75}) & 615 & 34.32\% & 447 (24.9\%) & \textbf{85.4\%} & 0.985 & 0.475 & 0.400 & 0.903 \\
\bottomrule
\end{tabular}%
}
\vspace{2pt}\par{\footnotesize All metrics reflect raw point clouds prior to Track~B valence reconciliation. \texttt{Blind Audit MAE} is scored by the isolated DFT-calibrated \texttt{DPA-4} electronic oracle after relaxation.}
\end{table}

\begin{figure}[!t]
    \centering
    \includegraphics[width=\textwidth]{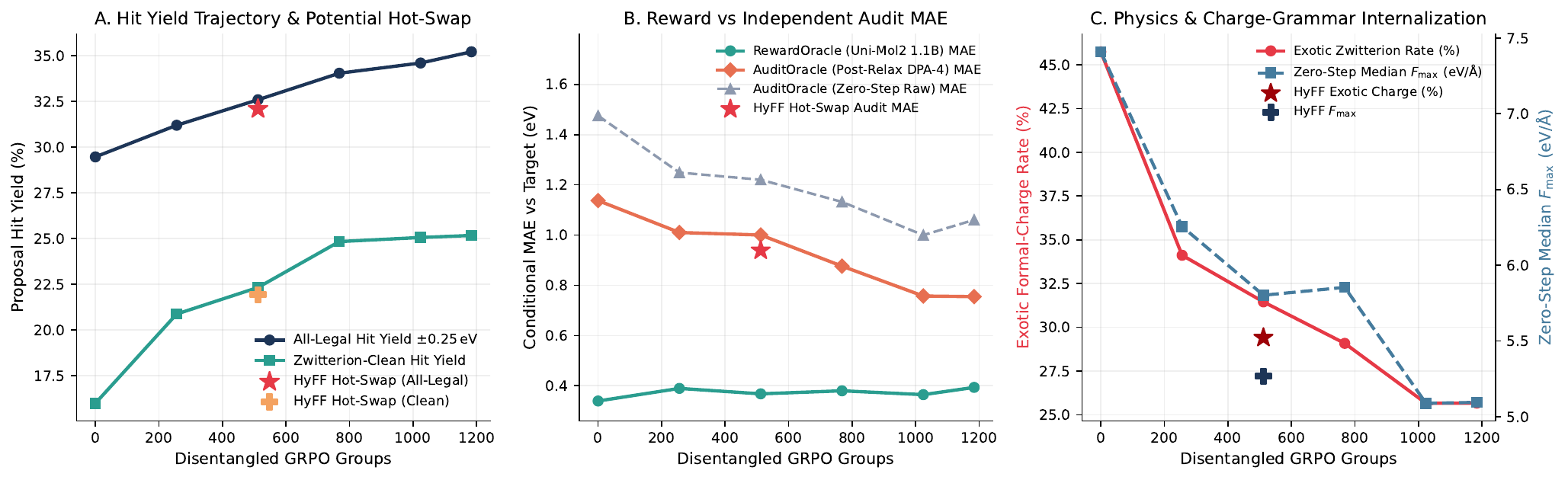}
    \caption{\textbf{Anti-reward-hacking trajectory across $0\text{--}1{,}184$ online \texttt{GRPO} groups and zero-code \texttt{PotentialSlot} hot-swapping.} (\textbf{A}) Total and clean hit yields vs.\ online RL budget. (\textbf{B}) Blind \texttt{AuditOracle} MAE vs.\ \texttt{RewardOracle} MAE. (\textbf{C}) Exotic formal-charge rate, zero-step median $F_{\max}$, and $512$-group hot-swap comparison between \texttt{DPA-4} and \texttt{HyFF vivace}.}
    \label{fig:audit_hotswap}
\end{figure}

\paragraph{Anti-Reward-Hacking Trajectory and Potential Hot-Swapping.}
Across $0 \to 1{,}184$ online \texttt{GRPO} groups evaluated every $256$ groups (Figure~\ref{fig:audit_hotswap}), isolated \texttt{AuditOracle} MAE decreases monotonically ($1.137 \to 1.010 \to 1.000 \to 0.876 \to 0.757 \to 0.755\,\text{eV}$) alongside falling zero-step median forces ($7.41 \to 5.09\,\text{eV/\AA}$) and rising MLIP graph-preservation rates ($75.8\% \to 90.6\%$), refuting Goodhart's-law exploitation. Furthermore, hot-swapping \texttt{PotentialSlot} from \texttt{DPA-4} to \texttt{HyFF vivace} \citep{hyra2026hyff} at $512$ groups via a single CLI flag requires zero model-code edits, matching hit yield ($32.09\%$, $575/1{,}792$) while improving initial atomic forces ($5.27$ vs.\ $5.80\,\text{eV/\AA}$), relaxation graph preservation ($88.3\%$ vs.\ $81.6\%$), and blind \texttt{Audit} MAE ($0.939$ vs.\ $1.000\,\text{eV}$).

\paragraph{Scale-Decoupling Ablation.}
Table~\ref{tab:v11_ablation} isolates why coarse topology RL and sub-voxel coordinate learning must remain decoupled. Propagating scalar electronic rewards through $3\times$ more numerous \texttt{XYZ} coordinate tokens alongside online relaxation distillation (\emph{Coupled RL}) yields strong local mechanics ($\text{RMSD}_{\text{Kabsch}} = 0.0495\,\text{\AA}$, $88.62\%$ legality, $F_{\max} = 4.19\,\text{eV/\AA}$) but inflates within-target electronic spread $\sigma_{\text{in}}$ from $0.439$ to $0.534\,\text{eV}$, cutting hit count from $620$ down to $526$. Restricting policy gradients to coarse tokens in Stage~2.5A preserves peak hit accuracy ($631$ hits, $0.744\,\text{eV}$ \texttt{Audit} MAE), while detached Stage~2.5B coordinate distillation lowers $\text{RMSD}_{\text{Kabsch}}$ to $0.0533\,\text{\AA}$ and raises raw legality to $85.38\%$.

\begin{table}[!t]
\centering
\caption{Ablation of spatial scale decoupling versus coupled coordinate reinforcement learning on \texttt{OE62 HOMO} ($1{,}792$ raw point-cloud proposals).}
\label{tab:v11_ablation}
\resizebox{\textwidth}{!}{%
\begin{tabular}{lrrrrrrr}
\toprule
Ablation Arm & Hits (Clean) & 3D Legal & Exotic $Q$ & $\sigma_{\text{in}}$ (eV) & Median $F_{\max}$ (eV/\AA) & $\text{RMSD}_{\text{Kabsch}}$ (\AA) & Post-Relax Audit MAE \\
\midrule
\texttt{v1.0} Baseline (no guard, frozen \texttt{XYZ}) & 620 (449) & 80.47\% & 25.7\% & 0.439 & 5.09 & 0.0613 & 0.757 \\
$+$ Zero-shot 3D geometric guard + tree \texttt{CFG} & 622 (444) & 81.75\% & 25.5\% & 0.436 & 5.12 & 0.0586 & 0.811 \\
$+$ Detached $0.01\,\text{\AA}$ \texttt{XYZ} equilibrium distillation & 625 (450) & 81.81\% & 25.8\% & \textbf{0.427} & 5.19 & 0.0581 & 0.825 \\
\textbf{Stage 2.5A} (\texttt{Guard + TopoGRPO}, coarse tokens only) & \textbf{631 (451)} & 84.60\% & 25.3\% & 0.471 & 5.10 & 0.0559 & \textbf{0.744} \\
\textbf{Stage 2.5A + 2.5B} (cascaded decoupled) & 615 (447) & 85.38\% & 26.1\% & 0.475 & 5.18 & 0.0533 & 0.903 \\
Control: \emph{Coupled RL} (\texttt{XYZ} in policy + online distill) & 526 (402) & \textbf{88.62\%} & \textbf{17.2\%} & 0.534 & \textbf{4.19} & \textbf{0.0495} & 0.786 \\
\bottomrule
\end{tabular}%
}
\vspace{2pt}\par{\footnotesize $\text{RMSD}_{\text{Kabsch}}$ measures all-atom conformational displacement after $20$-step \texttt{DPA-4} relaxation.}
\end{table}

\subsection{Multi-Property Reachability and Quantum Anchor Validation}
\label{sec:exp_atlas}

Conditioning a single unified checkpoint on $[\text{HOMO}, \text{LUMO}]$ with missingness masks supports both marginal and joint orbital design. On marginal \texttt{LUMO} generation ($1{,}792$ proposals), incorporating a $40$-step \texttt{DPA-4} equilibrium pool and 3D coordination guards lifts raw legality from $75.3\%$ to $87.33\%$, lowers exotic charges from $27.1\%$ to $16.2\%$, reduces median zero-step $F_{\max}$ from $5.04$ to $3.79\,\text{eV/\AA}$, achieves $94.1\%$ relaxation graph preservation ($\text{RMSD}_{\text{Kabsch}} = 0.0469\,\text{\AA}$), and reaches unit calibration slope ($1.002$).

\begin{figure}[!t]
    \centering
    \includegraphics[width=\textwidth]{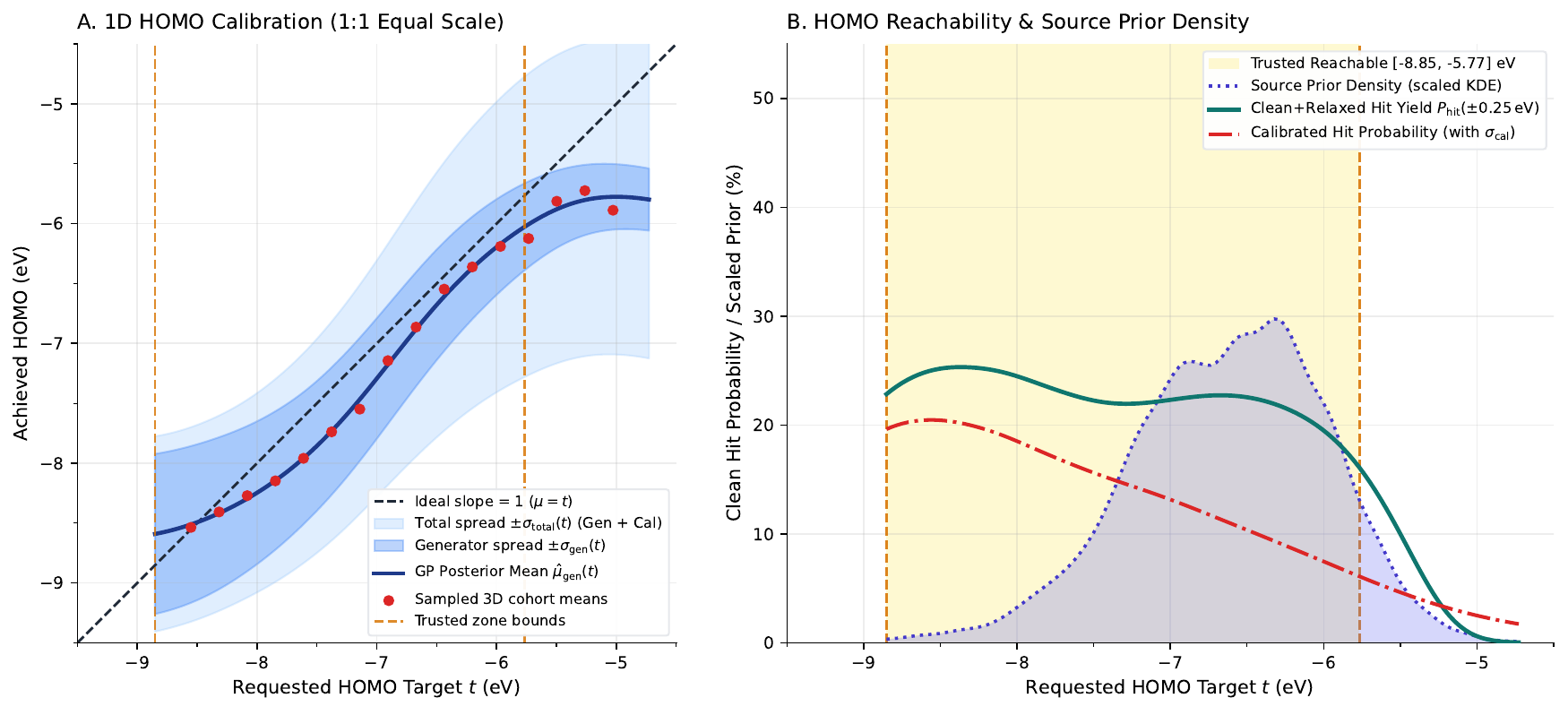}
    \caption{\textbf{Univariate 1D \texttt{HOMO} heteroscedastic \texttt{GPAtlas}.} (\textbf{A}) Equal-scale reachability calibration across $[-9.5, -4.5]\,\text{eV}$ showing achieved mean $\hat{\mu}_{\text{gen}}(t)$, generator spread $\pm\sigma_{\text{gen}}(t)$, and total calibrated spread $\pm\sigma_{\text{total}}(t)$. (\textbf{B}) Surrogate vs.\ calibrated hit probability across the trusted reachable interval $[-8.85, -5.77]\,\text{eV}$ juxtaposed against the \texttt{OE62} source prior density.}
    \label{fig:atlas1d}
\end{figure}

\begin{figure}[!t]
    \centering
    \includegraphics[width=\textwidth]{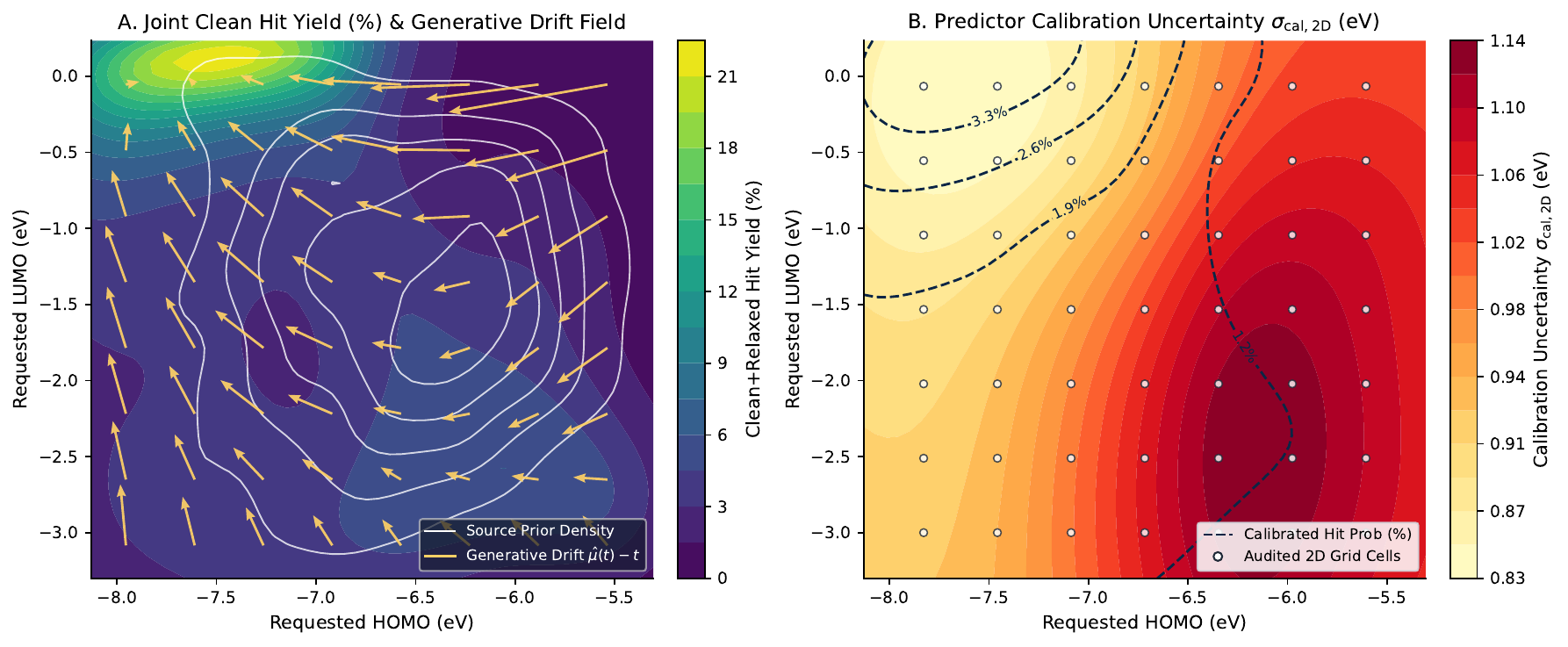}
    \caption{\textbf{Orthogonal 2D $(\text{HOMO}, \text{LUMO})$ reachability and cross-foundation uncertainty atlas ($49$ cells, $1{,}764$ clean molecules).} (\textbf{A}) Joint clean hit yield surface with \texttt{OE62} source density contours and generative drift vectors $\hat{\boldsymbol{\mu}}(\mathbf{t}) - \mathbf{t}$. (\textbf{B}) Empirical cross-foundation disagreement surface $\sigma_{\text{cal,2D}}(t_{\text{H}}, t_{\text{L}})$ between \texttt{Uni-Mol2} and calibrated \texttt{DPA-4}.}
    \label{fig:atlas2d}
\end{figure}

Figures~\ref{fig:atlas1d} and~\ref{fig:atlas2d} map the 1D and 2D heteroscedastic \texttt{GPAtlas} over $1{,}764$ relaxed, clean molecules across a $7 \times 7$ $(\text{HOMO}, \text{LUMO})$ grid. In 1D (Figure~\ref{fig:atlas1d}), posterior conditional mean $\hat{\mu}_{\text{gen}}(t)$ tracks the diagonal across the trusted reachable window $[-8.85, -5.77]\,\text{eV}$ before saturating against the \texttt{OE62} prior ceiling above $-5.77\,\text{eV}$. In 2D (Figure~\ref{fig:atlas2d}), joint clean yield peaks at $22.0\%$ with drift vectors pulling extreme queries toward dense semiconductor space, while empirical cross-foundation disagreement $\sigma_{\text{cal,2D}}$ rises smoothly from $0.83\,\text{eV}$ in data-rich wide-gap cells to $1.14\,\text{eV}$ in narrow-gap donor--acceptor corners.

\paragraph{Stratified Quantum Anchor Verification.}
To verify that cross-oracle disagreement hotspots in Figure~\ref{fig:atlas2d} reflect genuine quantum-mechanical surrogate breakdown rather than noise, we sampled $80$ unique \texttt{DPA-4}-relaxed clean molecules stratified across the $8$ lowest-divergence cells ($\text{cal.\ RMSE}_{2\text{D}} = 0.620\text{--}0.764\,\text{eV}$, $n=40$) and $8$ highest-divergence cells ($1.239\text{--}1.434\,\text{eV}$, $n=40$), running live \texttt{GPU4PySCF} \citep{sun2020pyscf,li2024gpu4pyscf} single-point \texttt{PBE0/def2-TZVP/DF} calculations alongside full \texttt{PBE-D3BJ/def2-TZVP/DF} geometry optimizations on $16$ cell representatives (Table~\ref{tab:atlas2d_dft}). Pre-stabilization with \texttt{DPA-4} enables $80/80$ ($100.0\%$) single-point DIIS convergence with zero Newton fallbacks and $16/16$ ($100.0\%$) full DFT relaxation convergence, preserving covalent bond graphs in $15/16$ ($93.75\%$) cases at $0.0540\,\text{\AA}$ median heavy-atom Kabsch RMSD. Against true \texttt{PBE0}, calibrated \texttt{DPA-4} tracks orbital energies tightly ($0.1893\,\text{eV}$ \texttt{HOMO} MAE, $r=0.9822$; $0.2879\,\text{eV}$ \texttt{LUMO} MAE, $r=0.9755$), whereas frozen \texttt{Uni-Mol2} drifts by $3\text{--}4\times$ ($0.7235 / 0.8677\,\text{eV}$). Moving from low-divergence to high-divergence atlas cells triggers a $+78.0\%$ jump in \texttt{Uni-Mol2} \texttt{PBE0} \texttt{HOMO} MAE ($0.5205 \to 0.9265\,\text{eV}$) and drops Spearman rank correlation from $0.7964$ to $0.4914$, confirming that \texttt{GPAtlas} pinpoints true quantum-chemical blindspots.

\begin{table*}[t]
\centering
\caption{Stratified 2D $(\text{HOMO},\text{LUMO})$ \texttt{GPAtlas} verification against live \texttt{PBE0/def2-TZVP/DF} single-point and \texttt{PBE-D3BJ/def2-TZVP/DF} geometry-optimization benchmarks.}
\label{tab:atlas2d_dft}
\small
\resizebox{\textwidth}{!}{%
\begin{tabular}{@{}llccccc@{}}
\toprule
\textbf{Cohort / Stratum} & \textbf{Oracle} & \textbf{2D MAE (eV)} & \textbf{HOMO MAE / RMSE} & \textbf{HOMO $r / \rho$} & \textbf{LUMO MAE / RMSE} & \textbf{LUMO $r / \rho$} \\
\midrule
Overall ($n=80$, SCF $100\%$) & DPA-4 & $0.2386$ & $0.1893 / 0.2497$ & $0.9822 / 0.9767$ & $0.2879 / 0.4005$ & $0.9755 / 0.9712$ \\
Emp.\ diff $0.8133\text{ eV}$ & Uni-Mol2 & $0.7956$ & $0.7235 / 1.0279$ & $0.7110 / 0.7292$ & $0.8677 / 1.1631$ & $0.7449 / 0.8120$ \\
\midrule
Low-Divergence ($n=40$) & DPA-4 & $0.2011$ & $0.1498 / 0.2038$ & $0.9798 / 0.9702$ & $0.2523 / 0.3542$ & $0.9787 / 0.9341$ \\
Emp.\ diff $0.7624\text{ eV}$ & Uni-Mol2 & $0.7639$ & $0.5205 / 0.7209$ & $0.7569 / 0.7964$ & $1.0073 / 1.3140$ & $0.6578 / 0.8139$ \\
\midrule
High-Divergence ($n=40$) & DPA-4 & $0.2761$ & $0.2288 / 0.2883$ & $0.9778 / 0.9728$ & $0.3235 / 0.4420$ & $0.9661 / 0.9722$ \\
Emp.\ diff $0.8641\text{ eV}$ & Uni-Mol2 & $0.8273$ & $0.9265 / 1.2624$ & $0.5573 / 0.4914$ & $0.7281 / 0.9895$ & $0.7751 / 0.7587$ \\
\midrule
\multicolumn{7}{@{}l@{}}{\textbf{Full $\text{PBE-D3BJ/def2-TZVP}$ Relaxation ($n=16$, $1/$cell):} $16/16$ ($100.0\%$) converged; $15/16$ ($93.75\%$) graph preserved;} \\
\multicolumn{7}{@{}l@{}}{heavy Kabsch RMSD $0.0540\text{ \AA}$ median ($0.1578\text{ \AA}$ mean); all-atom RMSD $0.1582\text{ \AA}$ median ($0.2698\text{ \AA}$ mean); adiabatic shift $0.3056 / 0.1785\text{ eV}$.} \\
\bottomrule
\end{tabular}%
}
\vspace{2pt}\par{\footnotesize Single-point calculations use \texttt{def2-universal-jkfit}, grid level $3$, and SCF convergence threshold $10^{-8}\,\text{Ha}$ ($5$ molecules per cell across $8$ low-divergence and $8$ high-divergence atlas cells).}
\end{table*}

\subsection{Synthesizable Design and Valence Grounding}
\label{sec:exp_v20_synth}

\begin{figure}[!t]
    \centering
    \includegraphics[width=\textwidth]{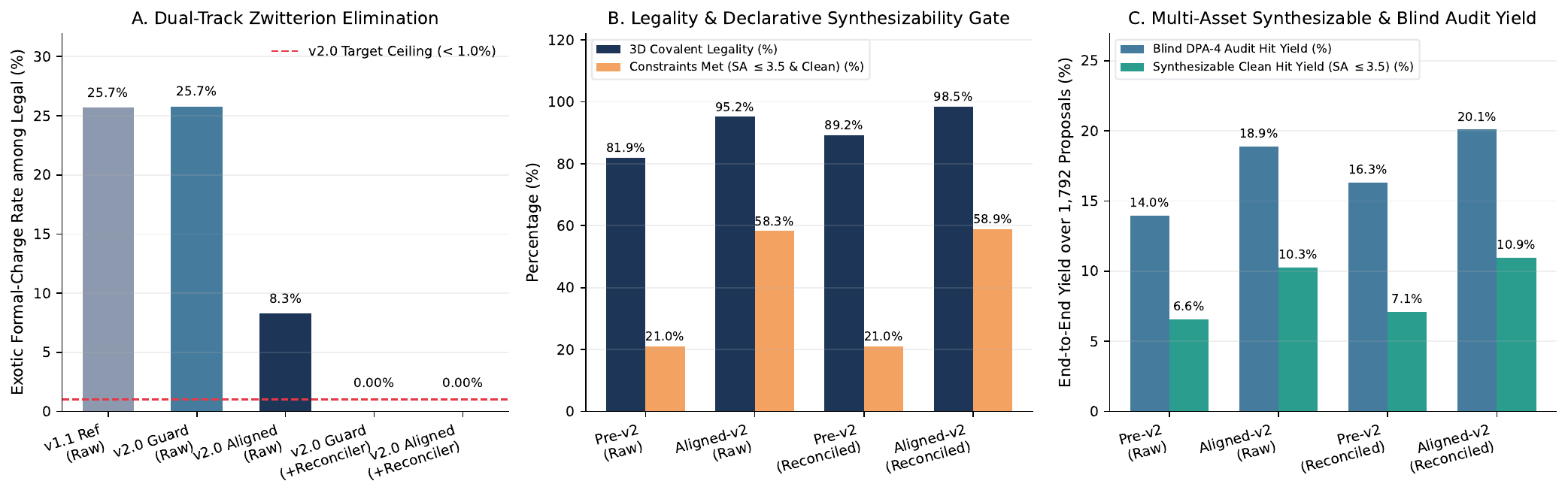}
    \caption{\textbf{Synthesizable multi-asset weaving and dual-track valence grounding ($1{,}792$ proposals).} (\textbf{A}) Exotic formal-charge rates before and after Track~A and Track~B grounding. (\textbf{B}) Covalent 3D legality and multi-constraint satisfaction ($\text{SA\_Score} \le 3.5$, $\text{SMARTS} = 0$, zero exotic charge). (\textbf{C}) Synthesizable clean hit yield, blind \texttt{DPA-4} audit hits, and collapse of the \texttt{Audit}--\texttt{Reward} MAE gap.}
    \label{fig:v20_synth}
\end{figure}

To test zero-retrain composition of frozen external predictors with medicinal/materials chemistry rules, we mounted frozen \texttt{Uni-Mol2} alongside declarative inequality constraints ($\text{SA\_Score} \le 3.5$ hard gate, reactive $\text{SMARTS} = 0$ hard gate, and $\text{MolWt} \le 550\,\text{Da}$ soft margin) and the isolated \texttt{DPA-4} blind auditor. Table~\ref{tab:v20_synth} and Figure~\ref{fig:v20_synth} compare \emph{Raw Point-Cloud} and \emph{Valence-Reconciled} views across $1{,}792$ proposals.

\begin{table}[!t]
\centering
\caption{Comparison of \emph{Raw Point-Cloud} and \emph{Valence-Reconciled} evaluation views under zero-retrain synthesizable weaving ($\text{SA\_Score} \le 3.5$ and $\text{SMARTS} = 0$) across $1{,}792$ \texttt{OE62 HOMO} proposals.}
\label{tab:v20_synth}
\resizebox{\textwidth}{!}{%
\begin{tabular}{llrrrrrrrr}
\toprule
Policy Checkpoint & Evaluation View & 3D Legal & Exotic $Q$ & Median SA ($\downarrow$) & Constr.\ Met & Median $F_{\max}$ & Synth.\ Clean Hits & Blind Audit Hits & Audit$-$Reward Gap \\
\midrule
Pre-Align \texttt{v1.1} & \textbf{Raw Point-Cloud} & 81.92\% (1468) & 25.75\% & 4.87 & 21.05\% & 5.07 & 118 (6.58\%) & 250 (13.95\%) & 0.655 eV \\
Pre-Align \texttt{v1.1} & \textbf{Valence-Reconciled} & 89.23\% (1599) & \textbf{0.00\%} & 4.70 & 20.95\% & 5.89 & 127 (7.09\%) & 292 (16.29\%) & 0.555 eV \\
\midrule
\textbf{Loom v2.0} & \textbf{Raw Point-Cloud} & \textbf{95.20\% (1706)} & \textbf{8.26\%} & \textbf{3.04} & \textbf{58.32\%} & \textbf{1.79} & \textbf{184 (10.27\%)} & \textbf{338 (18.86\%)} & \textbf{0.190 eV} \\
\textbf{Loom v2.0} & \textbf{Valence-Reconciled} & \textbf{98.49\% (1765)} & \textbf{0.00\%} & \textbf{3.04} & \textbf{58.87\%} & \textbf{1.91} & \textbf{196 (10.94\%)} & \textbf{360 (20.09\%)} & \textbf{0.143 eV} \\
\bottomrule
\end{tabular}%
}
\vspace{2pt}\par{\footnotesize \emph{Valence-Reconciled} locks all heavy-atom coordinates identically ($\max \text{RMSD}_{\text{heavy}} = 0.0000\,\text{\AA}$). \texttt{Audit$-$Reward Gap} measures $|\text{MAE}_{\text{Audit}} - \text{MAE}_{\text{Reward}}|$.}
\end{table}

Three findings stand out from Table~\ref{tab:v20_synth}. First, even without post-processing (\textbf{Raw Point-Cloud}), source-side Track~A raises 3D legality from $81.92\%$ to $95.20\%$, reduces disconnected skeletons from $110$ to $18$ and over-coordinated atoms from $66$ to $5$, lifts natively clean raw molecules from $1{,}090$ to $1{,}565$, and cuts raw exotic charges $3.1\times$ ($25.75\% \to 8.26\%$). Track~B then resolves the remaining $200$ boundary cases ($60$ peripheral-H count mismatches and $140$ neutral Lewis reassignments) at exact $0.0000\,\text{\AA}$ heavy-atom RMSD, reaching $98.49\%$ legality and $0.00\%$ exotic charges. Second, declarative synthesizability constraints suppress cross-foundation reward hacking: unconstrained \texttt{v1.1} exploited strained polycyclic cages ($\text{median SA} = 4.87$, $F_{\max} = 5.07\,\text{eV/\AA}$), incurring a $0.655\,\text{eV}$ \texttt{Audit}--\texttt{Reward} MAE gap. Enforcing $\text{SA\_Score} \le 3.5$ shifts generation onto unstrained conjugated scaffolds ($\text{median SA} = 3.04$, $F_{\max} = 1.79\text{--}1.91\,\text{eV/\AA}$), expanding synthesizable clean hits by $+66.1\%$ ($118 \to 196$, with $137$ unique and $120$ novel molecular graphs), lifting blind \texttt{DPA-4} hits by $+44.0\%$ ($250 \to 360$), and compressing the \texttt{Audit}--\texttt{Reward} gap $4.6\times$ to $0.143\,\text{eV}$. Third, restricting replay admission to $\text{SA} \le 3.5$ ($25.5\%$ of the unconstrained pool) preserves unit calibration slope ($1.028$) while exposing a genuine $-0.43\,\text{eV}$ chemical Pareto trade-off across sparse multi-objective conjugated regions \citep{wang2026mimflow}, as shallow-\texttt{HOMO} ($-5.85\,\text{eV}$) strong electron donors predominantly require fused heteroaromatic rings with $\text{SA} > 3.8$.

\FloatBarrier
\subsection{Closed-Loop Recursive Self-Improvement}
\label{sec:exp_rsi}

We next evaluate the four-quadrant RSI flywheel in a strict four-way segregated industrial blindspot protocol ($1{,}792$ proposals per round): (i) a $5$-fold enterprise QSAR ensemble wrapped in \texttt{ResidualCalibratedAsset} is trained exclusively on $3{,}086$ molecules with $\text{HOMO} > -6.92\,\text{eV}$, leaving deep electron acceptors ($\text{HOMO} \le -6.92\,\text{eV}$, covering $P_{10} = -7.45\,\text{eV}$) as a zero-data OOD blindspot guarded only by \texttt{single\_oracle\_ad} auditing and $\text{SA\_Score} \le 3.8$ ($\tau_{\text{cal}} = 0.35\,\text{eV}$); (ii) an isolated external \texttt{PBE0/def2-TZVP} pool ($73$ deep-\texttt{HOMO} structures) supplies authentic DFT labels only when queried by \texttt{OuterRSIAcquisition}; (iii) $73$ untouched deep-\texttt{HOMO} DFT molecules form a blind calibration test set; and (iv) the isolated \texttt{DPA-4 AuditOracle} scores all generated 3D proposals across mid-range ($P_{35}\text{--}P_{90}$) and deep-blindspot ($P_{10}\text{--}P_{25}$) quantiles.

\suppressfloats[t]
\begin{table}[ht]
\centering
\caption{Two-round closed-loop RSI progression across $1{,}792$ proposals per round under an enterprise QSAR blindspot ($\text{HOMO} \le -6.92\,\text{eV}$ withheld from base predictor training, $\tau_{\text{cal}} = 0.35\,\text{eV}$).}
\label{tab:rsi}
\resizebox{\textwidth}{!}{%
\begin{tabular}{llrrrrrrr}
\toprule
RSI Stage & Grid Zone Counts & $P_{10}$ $\sigma_{\text{cal}}$ & 3D Legal & Raw Exotic $Q$ & Constr.\ Met & Mid-Zone Audit & Deep-Zone Audit & Total Blind Audit \\
 & \texttt{(Conv/Inner/Blind/Ceil)} & ($-7.45$ eV) & (\texttt{Raw} / \texttt{Rec}) & (\texttt{Rec} $=0\%$) & ($\text{SA}\le 3.8$) & ($P_{35}\text{--}P_{90}$) & ($P_{10}\text{--}P_{25}$) & Clean Hits / 1792 \\
\midrule
\textbf{Round 0} (Initial Diagnosis) & \texttt{0 / 6 / 1 / 1} & 0.490 eV & 66.0\% / 76.1\% & 38.88\% & 13.27\% & 2.11\% & 4.88\% & \textbf{52 (2.90\%)} \\
\textbf{Round 1A} (Inner RSI Only, 0 DFT) & \texttt{3 / 3 / 1 / 1} & 0.483 eV & 94.9\% / 98.4\% & 7.35\% & 57.06\% & \textbf{8.59\%} ($4.07\times$) & 5.47\% & \textbf{138 (7.70\%, $2.65\times$)} \\
\textbf{Round 1B} (64 DFT + Zero-Retrain Cal) & \textbf{\texttt{2 / 5 / 0 / 1}} & \textbf{0.291 eV} & \multicolumn{5}{c}{\emph{Untouched Blind Deep-DFT Test ($n=73$) MAE: $0.990 \to 0.655$ eV (\textbf{$-33.8\%$}, $0$ base weights modified)}} & \emph{Unlocks $P_{10}\text{--}P_{25}$} \\
\textbf{Round 2} (Unlocked Inner RSI) & \textbf{\texttt{3 / 4 / 0 / 1}} & \textbf{0.285 eV} & \textbf{97.4\% / 99.3\%} & \textbf{3.61\%} & \textbf{75.04\%} & \textbf{14.69\%} ($6.96\times$) & \textbf{8.01\%} ($+64.1\%$) & \textbf{229 (12.78\%, $4.40\times$)} \\
\bottomrule
\end{tabular}%
}
\vspace{2pt}\par{\footnotesize Rightmost three columns report strictly blind \texttt{DPA-4 AuditOracle} hit rates on valence-clean, constraint-satisfying ($\text{SA\_Score} \le 3.8$) 3D molecules.}
\end{table}

\begin{figure}[!t]
    \centering
    \includegraphics[width=\textwidth]{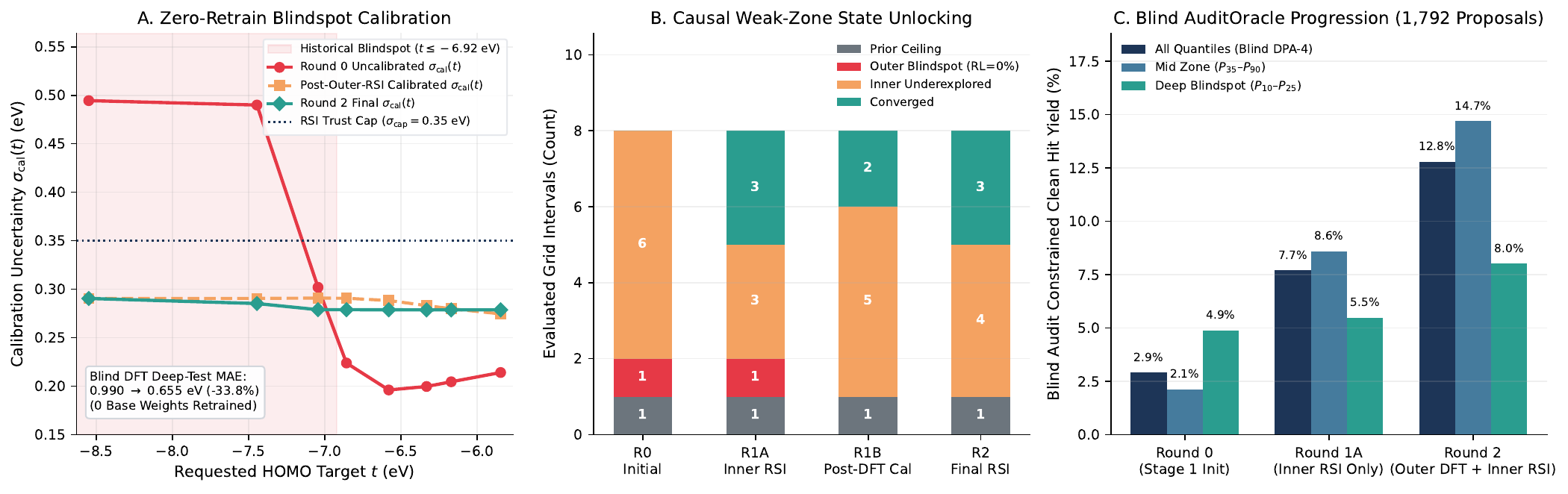}
    \caption{\textbf{Two-round physics-grounded RSI closed loop.} (\textbf{A}) Epistemic uncertainty $\sigma_{\text{cal}}(t)$ before and after zero-retrain residual calibration on $64$ outer-loop DFT acquisitions. (\textbf{B}) Diagnostic state counts across rounds showing blindspot protection in Round~1A and unlocking in Round~1B. (\textbf{C}) Blind \texttt{DPA-4 AuditOracle} constrained clean hit yields across mid-range and deep-blindspot targets.}
    \label{fig:rsi}
\end{figure}

Table~\ref{tab:rsi} and Figure~\ref{fig:rsi} verify every link of the causal self-improvement chain:
\begin{itemize}[leftmargin=1.5em,itemsep=2pt,topsep=2pt]
    \item \textbf{Round 0 (Diagnosis and Gatekeeper Routing):} \texttt{RSIDiagnoser} labels the extreme $-8.55\,\text{eV}$ probe as \texttt{PRIOR\_CEILING} ($\rho_{\text{src}} = 0.013$, pull bias $+2.06\,\text{eV}$) and $P_{10} = -7.45\,\text{eV}$ as \texttt{OUTER\_BLINDSPOT} ($\sigma_{\text{cal}} = 0.490\,\text{eV} > 0.35\,\text{eV}$), locking both to $0\%$ RL weight and routing \texttt{recommend\_target(-7.446)} to \texttt{rerouted\_warning} ($P_{\text{hit}}^{\text{cal}} = 5.6\%$).
    \item \textbf{Round 1A (Zero-DFT Inner Alignment):} Concentrating curriculum updates on the $6$ trusted \texttt{INNER\_UNDEREXPLORED} bins lifts mid-zone blind-audited clean hits $4.07\times$ ($2.11\% \to 8.59\%$), graduating $3$ bins to \texttt{CONVERGED} while leaving the frozen blindspot uncorrupted ($4.88\% \to 5.47\%$).
    \item \textbf{Round 1B (Outer Acquisition and Zero-Retrain Calibration):} \texttt{OuterRSIAcquisition} exports $64$ clean, low-strain candidates spanning $12$ Murcko scaffolds ($\bar{\sigma}_{\text{cal}} = 0.453\,\text{eV}$, uncalibrated $\text{MAE} = 1.187\,\text{eV}$). Updating \texttt{ResidualCalibratedAsset} with $0$ base-predictor weight edits cuts MAE on the untouched $n=73$ blind deep-DFT test set by $33.8\%$ ($0.990 \to 0.655\,\text{eV}$), contracts $\sigma_{\text{cal}}(-7.45\,\text{eV})$ to $0.291\,\text{eV}$, and clears \texttt{OUTER\_BLINDSPOT} ($1 \to 0$).
    \item \textbf{Round 2 (Unlocked Inner Alignment):} With deep acceptors unlocked, Round~2 boosts deep-zone blind-audited clean yield to $8.01\%$ ($+64.1\%$ over Round~0; deep \texttt{Audit} MAE $1.030 \to 0.955\,\text{eV}$), drives mid-zone yield to $14.69\%$ ($6.96\times$), and reaches $229 / 1{,}792$ ($12.78\%$, $4.40\times$) total blind clean hits at $97.38\%$ raw / $99.27\%$ reconciled legality, $3.61\%$ raw exotic charge, and $0.93\,\text{eV/\AA}$ median $F_{\max}$. Re-querying \texttt{recommend\_target(-7.446)} transitions automatically to \texttt{trusted} ($P_{\text{hit}}^{\text{cal}} = 44.85\%$).
\end{itemize}

\paragraph{Live \texttt{GPU4PySCF PBE0/def2-TZVP} Outer-Loop Validation.}
Replacing held-out pool lookup with on-the-fly quantum calculations directly on newly generated structures, we relaxed the $64$ \texttt{OuterRSIAcquisition} blindspot conformers for $25$ \texttt{DPA-4} steps and executed live \texttt{GPU4PySCF} \texttt{PBE0/def2-TZVP/DF} single points paired with uniform $20\%$ anchor replay across all previously trusted quantiles. All $64/64$ ($100.0\%$) generated conformers converged in $70.54\,\text{s}$ total ($0.926\,\text{s/mol}$ on one \texttt{A800} GPU; uncalibrated batch error was $1.257\,\text{eV}$). Zero-retrain residual calibration cuts untouched $n=73$ blind deep-DFT test MAE from $0.990$ to $0.823\,\text{eV}$ ($-16.9\%$) and clears \texttt{OUTER\_BLINDSPOT} ($1 \to 0$). Across Round~$0 \to \text{Round } 1\text{A} \to \text{Round } 2$, calibration slope recovers from $0.322 \to 0.332 \to 1.037$, deep-zone blind clean yield surges $8.17\times$ ($1.17\% \to 9.57\%$), total blind clean hits grow $9.65\times$ ($23 \to 151 \to 222 / 1{,}792$), and uniform trusted-quantile anchor replay cuts shallow-tail ($-5.85\,\text{eV}$) mean drift by $80\%$ ($-6.86 \to -6.055\,\text{eV}$).

\subsection{Multi-Property Conditional Generation on QM9}
\label{sec:exp_qm9}

To evaluate portability outside \texttt{OE62} and test generator-agnostic critic scaling, we benchmarked \textbf{Loom} initialized from unconditional \texttt{Uni-3DAR} (\texttt{qm9.pt}) \citep{lu2025uni3dar} alongside official \texttt{PropMolFlow} \citep{zeng2026propmolflow} checkpoints on the disjoint $50\text{k}/50\text{k}$ \texttt{QM9} split \citep{ramakrishnan2014qm9,zeng2026propmolflow} ($44{,}453$ train and $1{,}500$ dev conformers from \texttt{train\_rQM\_v0}, verified at $0$ overlap with the $50{,}000$-molecule \texttt{gvp\_rQM\_v0} evaluator split). All hindsight relabeling, RL rewards, dev sweeps, and test-time \texttt{Best-of-$M$} ($M\in\{1,2,4,8,16\}$) selections use exclusively an internal \texttt{Uni-Mol} \citep{zhou2023unimol} surrogate trained strictly on \texttt{train\_rQM\_v0} ($214.25\,\text{s}$), scoring only the final selected molecule per target ($N=1{,}000$/property) once on frozen \texttt{PropMolFlow} \texttt{GVPRegressor} oracles on a single \texttt{A800} GPU (Table~\ref{tab:qm9}).

We compare four \textbf{Loom} adaptation regimes under matched tree-annealed \texttt{CFG} and Track~A coordination guards: (i) one unified scalar-only $K=6$ SFT checkpoint ($31.60\text{M}$ trainable parameters, $0.968\,\text{GPU-h}$ total); (ii) six dedicated $K=1$ SFT checkpoints ($3.076\,\text{GPU-h}$ total); (iii) six $K=1$ Full checkpoints adding Hindsight and Stage~2.5A \texttt{TopoGRPO} ($4.500\,\text{GPU-h}$ total); and (iv) \textbf{Scheme~2}, six covariate-matched $(t, N_{\text{atoms}})$ checkpoints ($6.309\,\text{GPU-h}$ Full) in a canonical polarity-aligned inertial PCA frame matching \texttt{EDM}/\texttt{GeoLDM}/\texttt{PropMolFlow}'s $(t, N_{\text{atoms}})$ conditioning, alongside our upgraded full-$\text{SO}(3)$ \textbf{Loom [\texttt{MolGPAO}]} unified $K=6$ engine ($0.66\,\text{GPU-h}$ total compute \citep{chen2026molgpao}). Direct \texttt{A800} auditing of all six official \texttt{PropMolFlow} checkpoints ($155.97\,\text{GPU-h}$ total) confirms $28.73\,\text{mol/s}$ single-shot throughput ($100$ ODE steps), $95.0\%\text{--}98.9\%$ native sanitization, and $96.1\%\text{--}98.5\%$ coordinate-reconciled legality with $0.00\%$ exotic charges.

Three findings emerge across single-shot ($M=1$) and critic-guided ($M=1\to 16$) regimes:
\begin{enumerate}[leftmargin=1.5em,itemsep=2pt,topsep=2pt]
    \item \textbf{Covariate Parity and Polarity Alignment ($M=1$):} Moving from $1\times K=6$ SFT to $6\times K=1$ SFT lowers single-shot raw MAE on $5/6$ tasks ($\alpha: 5.43 \to 4.33\,\text{Bohr}^3$), while \textbf{Scheme~2} shrinks atom-count error $3\times$ ($|\Delta N|: 3.12 \to 0.98$ raw), cuts extensive $C_v$ MAE by $-23.2\%$ ($3.114 \to 2.390$ reconciled), and lifts single-shot dipole $\mu$ correlation $3.9\times$ ($r: 0.076 \to 0.298$) by removing rotational frame smearing, though single-pass discrete octree sampling ($80.3\text{--}92.6\,\text{mol/s}$) retains wider single-sample spread than $100$-step continuous ODE integration.
    \item \textbf{High-Throughput Critic-Guided Weaving ($M=1 \to 16$):} Activating \textbf{Loom}'s zero-leakage \texttt{CriticBus} and \texttt{ValenceReconciler} cuts error by $55\%\text{--}70\%$ with $100.0\%$ closed-shell legality and $0.00\%$ exotic charges at $M\ge 8$. With zero retraining on the $0.97\,\text{GPU-h}$ unified $1\times K=6$ checkpoint (\textbf{Scheme~1}), $M=4$ ($21.44\,\text{mol/s}$, matching \texttt{PropMolFlow} $M=1$ wall-clock speed) surpasses \texttt{EDM} and \texttt{GeoLDM} on $\Delta\varepsilon$, $\varepsilon_{\text{HOMO}}$, $\varepsilon_{\text{LUMO}}$, and $\mu$; at $M=16$ ($5.36\,\text{mol/s}$, $161\times$ less training compute than \texttt{PropMolFlow}), it outperforms single-shot \texttt{PropMolFlow} on $\Delta\varepsilon$ ($374.63\,\text{meV}$), $\varepsilon_{\text{HOMO}}$ ($199.19\,\text{meV}$), and $\mu$ ($0.4740\,\text{D}$). \textbf{Scheme~2} ($6.31\,\text{GPU-h}$) further lowers $\Delta\varepsilon$ to $358.16\,\text{meV}$ and cuts extensive $C_v$ MAE by $-31.2\%$ to $1.2619\,\text{cal/mol}\cdot\text{K}$.
    \item \textbf{Generator-Agnostic Synergy and Full-$\text{SO}(3)$ \texttt{MolGPAO} Upgrade:} Mounting the exact same zero-leakage \texttt{Uni-Mol} critic and \texttt{ValenceReconciler} onto \texttt{PropMolFlow} cuts its own errors by $59\%\text{--}75\%$ across $M=1 \to 16$, driving every property to the \texttt{GVP} holdout noise floor ($r=0.968\text{--}0.998$) at $100.0\%$ legality. Replacing raw \texttt{Uni-3DAR} with full-$\text{SO}(3)$ \textbf{MolGPAO} ($0.66\,\text{GPU-h}$, $235\times$ less compute than \texttt{PropMolFlow}) accelerates $M=1$ output to $121.22\,\text{mol/s}$ ($4.2\times$ faster than \texttt{PropMolFlow}) and reaches $\mu = 0.4171\,\text{D}$, $\varepsilon_{\text{LUMO}} = 329.21\,\text{meV}$, $\varepsilon_{\text{HOMO}} = 206.61\,\text{meV}$, and $\Delta\varepsilon = 390.09\,\text{meV}$ at $M=16$ with $100.0\%$ legality and $0.00\%$ exotic charges.
\end{enumerate}

\begin{table*}[t]
\centering
\caption{Six-property conditional generation and zero-leakage \texttt{Best-of-$M$} benchmark on the disjoint $50\text{k}/50\text{k}$ \texttt{QM9} split ($N=1{,}000$ targets per property, single \texttt{NVIDIA A800-80GB}).}
\label{tab:qm9}
\small
\setlength{\tabcolsep}{3.0pt}
\resizebox{\textwidth}{!}{%
\begin{tabular}{@{}lccrrrrrrrrr@{}}
\toprule
\textbf{Method / Weaving Protocol} & \textbf{\#} & \textbf{$M$} & \textbf{$\alpha$} & \textbf{$\Delta\varepsilon$} & \textbf{$\varepsilon_{\text{HOMO}}$} & \textbf{$\varepsilon_{\text{LUMO}}$} & \textbf{$\mu$} & \textbf{$C_v$} & \textbf{Legal / Exotic $Q$} & \textbf{Train} & \textbf{Eff.\ Out} \\
 & & & ($\text{Bohr}^3$) & (meV) & (meV) & (meV) & (D) & ($\text{cal/mol}\cdot\text{K}$) & (\%) & (mol/s $\mid$ GPU-h) & (mol/s) \\
\midrule
GVP Oracle Lower-Bound ($n=973$) & --- & --- & $0.5544$ & $190.08$ & $83.78$ & $135.89$ & $0.1422$ & $0.1726$ & $94.9\to97.3$ / --- & --- & --- \\
Random Target Baseline & --- & --- & $9.3116$ & $1423.78$ & $653.09$ & $1481.53$ & $1.5177$ & $4.5040$ & --- & --- & --- \\
\midrule
EDM (published)~\cite{hoogeboom2022edm} & 6 & 1 & $2.76$ & $655.0$ & $356.0$ & $584.0$ & $1.111$ & $1.101$ & ---$^\dagger$ & --- & --- \\
GeoLDM (published)~\cite{xu2023geoldm} & 6 & 1 & $2.37$ & $587.0$ & $340.0$ & $522.0$ & $1.108$ & $1.025$ & ---$^\dagger$ & --- & --- \\
PropMolFlow (published)~\cite{zeng2026propmolflow} & 6 & 1 & $1.3115$ & $390.98$ & $253.99$ & $314.52$ & $0.6202$ & $0.6259$ & ---$^\dagger$ & --- & --- \\
PropMolFlow (A800 Native audit) & 6 & 1 & $1.2688$ & $367.22$ & $256.43$ & $322.82$ & $0.6189$ & $0.6004$ & $95.0\text{--}98.9$ / $\le1.11$ & $996.25 \mid 155.97$ & $28.73$ \\
PropMolFlow (Coord Reconciled) & 6 & 1 & $1.4414$ & $410.57$ & $262.17$ & $336.61$ & $0.6526$ & $0.6259$ & $96.1\text{--}98.5$ / \textbf{0.00} & $996.25 \mid 155.97$ & $28.11$ \\
PropMolFlow + Loom Critic (Native) & 6 & 4 & $0.6511$ & $188.43$ & $114.86$ & $143.28$ & $0.2498$ & $0.2713$ & \textbf{100.0} / $\le0.10$ & $996.25 \mid 155.97$ & $7.22$ \\
PropMolFlow + Loom Critic (Native) & 6 & 8 & $0.5148$ & $162.75$ & $91.08$ & $121.70$ & $0.1855$ & $0.2092$ & \textbf{100.0} / \textbf{0.00} & $996.25 \mid 155.97$ & 3.60 \\
PropMolFlow + Loom Critic (Native) & 6 & 16 & \textbf{0.4566} & \textbf{145.92} & \textbf{79.42} & \textbf{113.40} & \textbf{0.1587} & \textbf{0.1771} & \textbf{100.0} / \textbf{0.00} & $996.25 \mid 155.97$ & 1.80 \\
PropMolFlow + Loom Critic (Coord-Rec) & 6 & 16 & $0.5323$ & $163.03$ & $81.56$ & $122.82$ & $0.1630$ & $0.2083$ & \textbf{100.0} / \textbf{0.00} & $996.25 \mid 155.97$ & 1.75 \\
\midrule
\textbf{Loom ($K=6$ SFT, Raw / Rec)} & 1 & 1 & $5.43 / 5.66$ & $1089 / 1074$ & $571.3 / 532.1$ & $1100 / 1054$ & $1.719 / 1.590$ & $2.933 / 3.114$ & $85.7\text{--}89.1$ / \textbf{0.00} & $220.27 \mid \mathbf{0.97}$ & $84.70$ \\
\textbf{Loom ($K=1$ SFT, Raw / Rec)} & 6 & 1 & $4.33 / 4.90$ & $1147 / 1094$ & $522.2 / 509.0$ & $1073 / 1063$ & $1.644 / 1.523$ & $2.460 / 2.838$ & $74.0\text{--}86.0$ / \textbf{0.00} & $194.18 \mid 3.08$ & $60.21$ \\
\textbf{Loom ($K=1$ Full, Raw / Rec)} & 6 & 1 & $5.28 / 5.50$ & $1318 / 1198$ & $597.9 / 527.3$ & $1105 / 1093$ & $1.815 / 1.809$ & $2.669 / 2.866$ & $80.6\text{--}95.8$ / \textbf{0.00} & $195.79 \mid 4.50$ & \textbf{92.63} \\
\textbf{Loom S2 ($[t,N]$ PCA, Raw / Rec)} & 6 & 1 & $4.91 / 5.15$ & $1037 / 1085$ & $516.7 / 507.1$ & $1118 / 1093$ & $1.624 / 1.514$ & $2.064 / 2.390$ & $84.4\text{--}91.1$ / \textbf{0.00} & $207.52 \mid 6.31$ & $80.33$ \\
\midrule
\textbf{Loom S1 ($K=6$ SFT, Rec)} & 1 & 4 & $3.6900$ & $602.05$ & $289.92$ & $543.32$ & $0.7218$ & $2.1888$ & $99.5\text{--}100$ / \textbf{0.00} & $220.27 \mid \mathbf{0.97}$ & $21.44$ \\
\textbf{Loom S1 ($K=6$ SFT, Rec)} & 1 & 8 & $3.4664$ & $449.25$ & $221.65$ & $407.04$ & $0.5601$ & $2.0075$ & \textbf{100.0} / \textbf{0.00} & $220.27 \mid \mathbf{0.97}$ & $10.71$ \\
\textbf{Loom S1 ($K=6$ SFT, Rec, $t$-only)} & 1 & 16 & $3.0801$ & $374.63$ & $199.19$ & $341.31$ & $0.4740$ & $1.8972$ & \textbf{100.0} / \textbf{0.00} & $220.27 \mid \mathbf{0.97}$ & $5.36$ \\
\textbf{Loom S1 ($K=6$ SFT, Rec, $t+N$)} & 1 & 16 & $2.9810$ & $388.67$ & $205.64$ & $364.78$ & $0.5051$ & $1.8332$ & \textbf{100.0} / \textbf{0.00} & $220.27 \mid \mathbf{0.97}$ & $5.36$ \\
\textbf{Loom S2 ($[t,N]$ PCA, Rec)} & 6 & 4 & $3.7355$ & $574.99$ & $285.56$ & $573.43$ & $0.8027$ & $1.5811$ & $99.1\text{--}100$ / \textbf{0.00} & $207.52 \mid 6.31$ & $21.00$ \\
\textbf{Loom S2 ($[t,N]$ PCA, Rec)} & 6 & 8 & $3.2677$ & $431.34$ & $229.50$ & $442.24$ & $0.6303$ & $1.3643$ & $99.9\text{--}100$ / \textbf{0.00} & $207.52 \mid 6.31$ & $10.63$ \\
\textbf{Loom S2 ($[t,N]$ PCA, Raw)} & 6 & 16 & $2.7457$ & $348.85$ & $216.30$ & $389.34$ & $0.6216$ & $1.1496$ & $99.5\text{--}100$ / $\le0.40$ & $207.52 \mid 6.31$ & $5.31$ \\
\textbf{Loom S2 ($[t,N]$ PCA, Rec)} & 6 & 16 & $3.0671$ & $358.16$ & $204.34$ & $380.21$ & $0.5434$ & $1.2619$ & \textbf{100.0} / \textbf{0.00} & $207.52 \mid 6.31$ & $5.31$ \\
\midrule
\textbf{Loom [MolGPAO ($K=6$)]} & 1 & 1 & $6.2547$ & $1201.4$ & $525.26$ & $1192.0$ & $1.5491$ & $3.6233$ & $86.5\text{--}89.5$ / \textbf{0.00} & $231.7 \mid \mathbf{0.66}$ & \textbf{121.22} \\
\textbf{Loom [MolGPAO ($K=6$)]} & 1 & 4 & $4.8611$ & $525.75$ & $256.24$ & $450.55$ & $0.6356$ & $2.7540$ & \textbf{100.0} / \textbf{0.00} & $231.7 \mid \mathbf{0.66}$ & $15.37$ \\
\textbf{Loom [MolGPAO ($K=6$)]} & 1 & 8 & $4.7729$ & $411.03$ & $218.51$ & $350.98$ & $0.5207$ & $2.5237$ & \textbf{100.0} / \textbf{0.00} & $231.7 \mid \mathbf{0.66}$ & $8.13$ \\
\textbf{Loom [MolGPAO ($K=6$)]} & 1 & 16 & $4.5797$ & $390.09$ & $206.61$ & $329.21$ & $0.4171$ & $2.2567$ & \textbf{100.0} / \textbf{0.00} & $231.7 \mid \mathbf{0.66}$ & $4.13$ \\
\bottomrule
\end{tabular}%
}
\vspace{3pt}\par{\footnotesize
All $M>1$ selections rank candidates exclusively with the \texttt{train\_rQM\_v0} \texttt{Uni-Mol} surrogate ($0$ overlap with \texttt{gvp\_rQM\_v0}) before single Top-$1$ evaluation on frozen \texttt{GVPRegressor} oracles; \texttt{Eff.\ Out} includes reconciliation and surrogate scoring at steady-state batch $250$ (\texttt{Loom}) or max batch $128$ (\texttt{PropMolFlow}). Single-shot $M=1$ ``Raw / Rec'' rows display reconciled legality ($0.00\%$ exotic $Q$); raw $M=1$ legality is $64.4\text{--}69.6\%$ ($K=6$), $50.2\text{--}64.4\%$ ($K=1$ SFT), $56.5\text{--}87.8\%$ ($K=1$ Full), and $67.5\text{--}78.8\%$ (S2). $^\dagger$Published EDM/GeoLDM tables omit RDKit sanitization, formal charges, and A800 timings.}
\end{table*}

\subsection{Cross-Paradigm Scaling on OE62}
\label{sec:exp_oe62_scaling}

Unlike small $\le 29$-atom \texttt{QM9} organics, \texttt{OE62} semiconductors reach $80$ atoms across $12$ elements under joint electronic ($|\hat{y}_{\text{HOMO}} - t| \le 0.25\,\text{eV}$) and synthetic accessibility ($\text{SA\_Score} \le 3.5$, $\text{SMARTS} = 0$) constraints. When candidate atom counts are resampled freely across $M$ draws, \texttt{Best-of-$M$} cherry-picks smaller scaffolds ($37.5 \to 30.0$ mean atoms at $M=8$) that satisfy $\text{SA}\le 3.5$ more easily. To eliminate size confounds and platform-calibration asymmetries on uncontended \texttt{A800-80GB} GPUs with zero inference-time MLIP calls, we held every prompt's $(t_i, N_{\text{atoms},i})$ pair ($N = 1{,}792$ prompts, $N_{\text{atoms},i}\in[18,42]$, mean $30.38$ atoms) strictly fixed across $M\in\{1,2,4,8,12,16\}$ and enforced identical $112$-dev affine calibration, $29$-D \texttt{ResidualCalibratedAsset}, Bayesian box-hit selection, blind \texttt{audit\_dpa4} scoring, and $42$-prompt live \texttt{PBE0/def2-TZVP/DF} verification across all \textbf{Loom} arms.

\begin{figure}[!t]
    \centering
    \includegraphics[width=\textwidth]{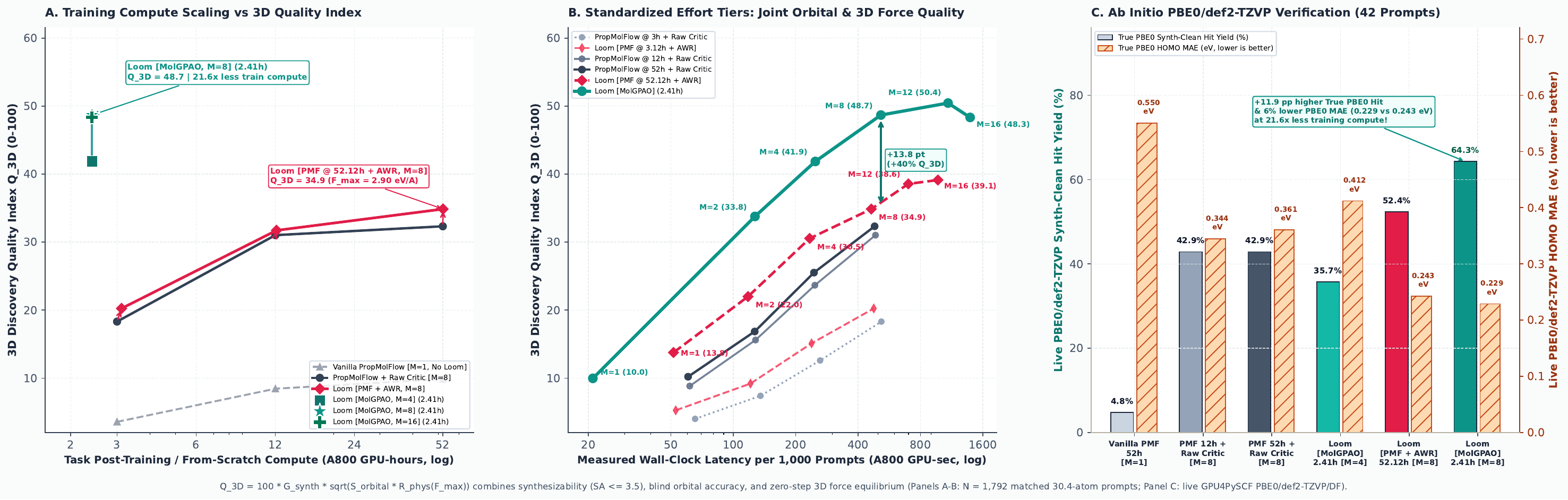}
    \caption{\textbf{Cross-paradigm \texttt{LoomService} scaling ($M\in\{1,2,4,8,12,16\}$) and live \texttt{PBE0/def2-TZVP} verification on size-matched \texttt{OE62} prompts ($N=1{,}792$, mean $30.4$ atoms, zero inference-time MLIP).} (\textbf{A}) Discovery quality $Q_{\text{3D}}$ vs.\ training compute ($2.41\text{--}52.12\,\text{GPU-h}$). (\textbf{B}) Latency vs.\ $Q_{\text{3D}}$ across service tiers $M=1\text{--}16$. (\textbf{C}) Unbiased $42$-prompt \texttt{GPU4PySCF PBE0/def2-TZVP} hit yield and \texttt{HOMO} MAE.}
    \label{fig:oe62_scaling}
\end{figure}

\begin{figure}[!t]
    \centering
    \includegraphics[width=\textwidth]{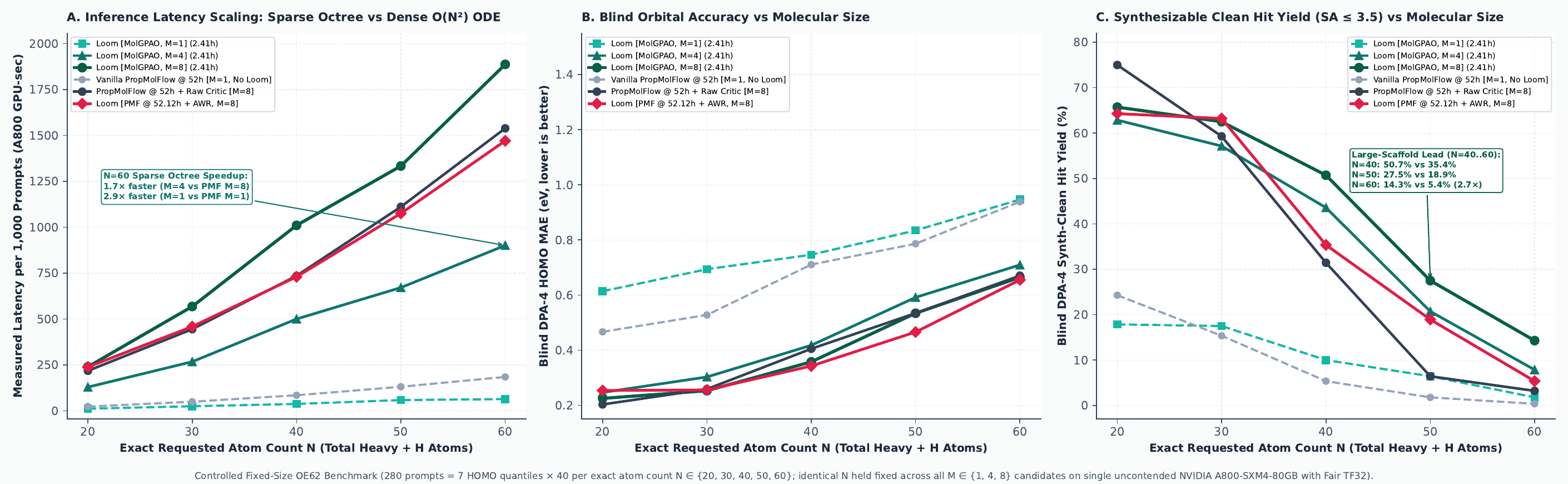}
    \caption{\textbf{Controlled molecular-size scaling across exact atom budgets $N\in\{20,30,40,50,60\}$ on \texttt{OE62} ($280$ fixed-size prompts per bucket).} (\textbf{A}) Latency per $1{,}000$ prompts comparing sparse octree decoding against dense $\mathcal{O}(N^2)$ flow matching. (\textbf{B}) Blind \texttt{DPA-4} \texttt{HOMO} MAE vs.\ $N$. (\textbf{C}) Blind synthesizable clean hit yield ($\text{SA\_Score}\le 3.5$) vs.\ $N$.}
    \label{fig:oe62_atom_scaling}
\end{figure}

\begin{figure}[!t]
    \centering
    \includegraphics[width=\textwidth]{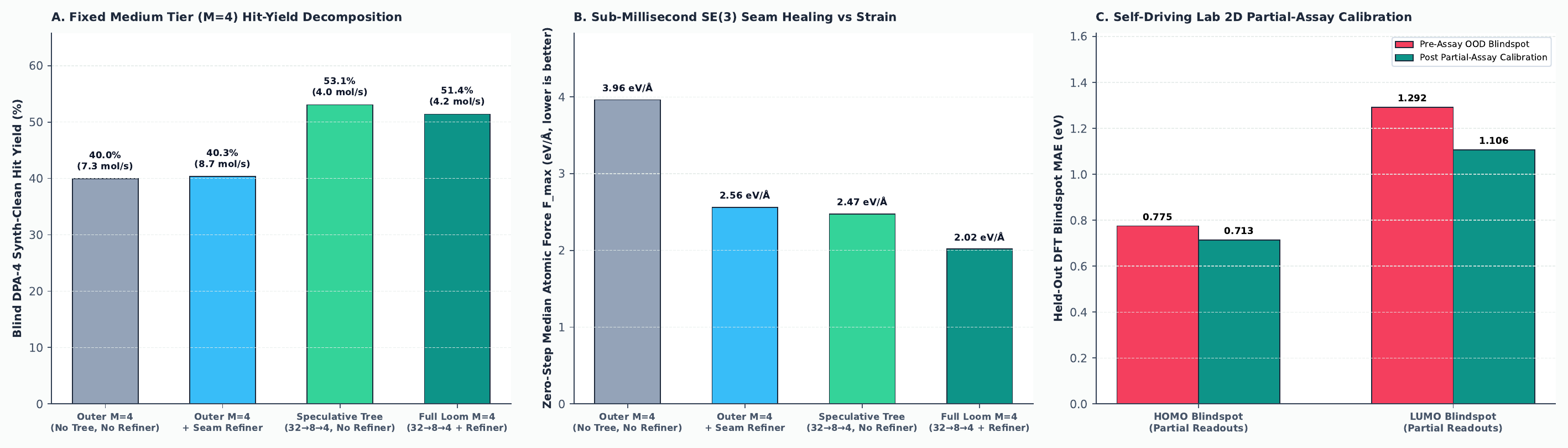}
    \caption{\textbf{Internal $M=4$ mechanism ablation and asynchronous 2D partial-assay calibration on \texttt{OE62}.} (\textbf{A}) Two-stage speculative octree pruning ($32\to 8\to 4$) vs.\ outer-only pruning ($4\to 4\to 4$) at fixed $M=4$ surrogate evaluations. (\textbf{B}) Zero-step atomic force $F_{\max}$ reduction from sub-millisecond \texttt{PostTreeGraphRefiner} without MLIP relaxation. (\textbf{C}) Held-out blindspot DFT MAE reduction from asynchronous 2D $(\text{HOMO},\text{LUMO})$ partial-\texttt{NaN} assay ingestion.}
    \label{fig:oe62_ablation_sdl}
\end{figure}

We compare pre-onboarding \texttt{PropMolFlow + Raw Critic} \citep{zeng2026propmolflow} ($3.0 \to 12.0 \to 52.0\,\text{GPU-h}$), onboarded \textbf{Loom [\texttt{PMF + AWR}]} ($3.12$, $12.12$, and $52.12\,\text{GPU-h}$, adding $+0.12\,\text{GPU-h}$ online \texttt{AWR}), and full-$\text{SO}(3)$ \textbf{Loom [\texttt{MolGPAO}]} ($2.41\,\text{GPU-h}$ total compute: $1.16\,\text{h}$ rank-$64$ \texttt{LoRA} SFT, $0.15\,\text{h}$ Hindsight \texttt{AWR}, $0.55\,\text{h}$ \texttt{TopoGRPO}, $0.03\,\text{h}$ detached \texttt{XYZ} distillation, $0.06\,\text{h}$ \texttt{PostTreeGraphRefiner}, and $0.46\,\text{h}$ \texttt{TwoStageOctreeProcessCritic}). To summarize synthesizability, blind orbital accuracy, and zero-step mechanical quality without arbitrary weights, Figures~\ref{fig:oe62_scaling}--\ref{fig:oe62_ablation_sdl} and Table~\ref{tab:oe62_scaling} report the 3D Discovery Quality Index $Q_{\text{3D}} = 100\,G_{\text{synth}}\sqrt{S_{\text{orb}}\,R_{\text{phys}}(F_{\max})} \in [0, 100]$.

\begin{table*}[t]
\centering
\caption{Symmetric \texttt{LoomService} ($M\in\{1,2,4,8,12,16\}$) cross-paradigm benchmark on fixed-size $30.4$-atom \texttt{OE62} prompts ($N=1{,}792$, $\text{SA\_Score}\le 3.5$, $\text{SMARTS}=0$, zero inference-time MLIP) and $42$-prompt live \texttt{PBE0/def2-TZVP} verification.}
\label{tab:oe62_scaling}
\small
\setlength{\tabcolsep}{2.3pt}
\resizebox{\textwidth}{!}{%
\begin{tabular}{@{}llrrrrrrrrrrr@{}}
\toprule
\textbf{Paradigm / Weaving Arm} & \textbf{Service} & \textbf{Train} & \textbf{Out} & \textbf{50k} & \textbf{Atoms /} & \textbf{SA$\le$3.5 /} & \textbf{$F_{\max}$} & \textbf{Surr.\ Synth} & \textbf{Blind DPA-4} & \textbf{Blind} & \textbf{$Q_{\text{3D}}$} & \textbf{Unbiased PBE0} \\
 & \textbf{Tier ($M$)} & \textbf{(GPU-h)} & \textbf{(mol/s)} & \textbf{(GPU-h)} & \textbf{Scaffolds} & \textbf{Legal (\%)} & \textbf{(eV/\AA)} & \textbf{Hits (\%)} & \textbf{Hits (\%)} & \textbf{MAE} & \textbf{(0--100)} & \textbf{Hit (Raw / Cen)} \\
\midrule
\textbf{Loom [MolGPAO]} & \texttt{base ($M=1$)} & \textbf{2.41} & \textbf{47.58} & \textbf{2.70} & $27.8$ / $1280$ & $56.1$ / $89.4$ & $3.82$ & $331$ ($18.5\%$) & $302$ ($16.9\%$) & $0.678$ & $10.0$ & --- \\
\textbf{Loom [MolGPAO]} & \texttt{low ($M=2$)} & \textbf{2.41} & $7.86$ & $4.18$ & $29.6$ / \textbf{1284} & $90.0$ / $99.6$ & $2.19$ & $914$ ($51.0\%$) & $745$ ($41.6\%$) & $0.418$ & $33.8$ & --- \\
\textbf{Loom [MolGPAO]} & \texttt{med ($M=4$)} & \textbf{2.41} & $4.02$ & $5.86$ & $29.6$ / $1226$ & $96.9$ / \textbf{100} & $2.06$ & $1305$ ($72.8\%$) & $920$ ($51.3\%$) & $0.343$ & $41.9$ & $35.7\%$ / $21.4\%$ \\
\textbf{Loom [MolGPAO]} & \texttt{high ($M=8$)} & \textbf{2.41} & $1.94$ & $9.57$ & $29.6$ / $1207$ & $99.8$ / \textbf{100} & \textbf{1.91} & $1653$ ($92.2\%$) & $1065$ ($59.4\%$) & $0.295$ & $48.7$ & $64.3\%$ / $64.3\%$ \\
\textbf{Loom [MolGPAO]} & \texttt{ext ($M=12$)} & \textbf{2.41} & $0.92$ & $17.51$ & $29.6$ / $1168$ & $99.9$ / \textbf{100} & $1.93$ & $1725$ ($96.3\%$) & \textbf{1117} (\textbf{62.3\%}) & $0.275$ & \textbf{50.4} & --- \\
\textbf{Loom [MolGPAO]} & \texttt{xhigh ($M=16$)} & \textbf{2.41} & $0.72$ & $21.70$ & $29.6$ / $1202$ & \textbf{100} / \textbf{100} & $2.01$ & \textbf{1756} (\textbf{98.0\%}) & $1081$ ($60.3\%$) & $0.286$ & $48.3$ & \textbf{66.7\%} / \textbf{69.0\%} \\
\midrule
\texttt{PropMolFlow + Raw Critic} & \texttt{high ($M=8$)} & $3.00$ & $1.93$ & $10.20$ & $30.4$ / $1257$ & $83.3$ / \textbf{100} & $3.72$ & $940$ ($52.5\%$) & $553$ ($30.9\%$) & $0.598$ & $18.3$ & --- \\
\texttt{PropMolFlow + Raw Critic} & \texttt{high ($M=8$)} & $12.00$ & $2.06$ & $18.74$ & $30.5$ / $1270$ & $91.2$ / \textbf{100} & $3.00$ & $1270$ ($70.9\%$) & $867$ ($48.4\%$) & $0.346$ & $31.0$ & $42.9\%$ / $47.6\%$ \\
\texttt{Vanilla PropMolFlow} (No Loom) & \texttt{base ($M=1$)} & $52.00$ & $21.20$ & $52.66$ & $30.1$ / $1221$ & $42.4$ / $89.8$ & $3.09$ & $347$ ($19.4\%$) & $273$ ($15.2\%$) & $0.512$ & $10.2$ & $4.8\%$ / $16.7\%$ \\
\texttt{PropMolFlow + Raw Critic} & \texttt{high ($M=8$)} & $52.00$ & $2.08$ & $58.68$ & $30.4$ / $1187$ & $88.5$ / \textbf{100} & $3.05$ & $1339$ ($74.7\%$) & $948$ ($52.9\%$) & $0.298$ & $32.3$ & $42.9\%$ / $54.8\%$ \\
\midrule
\textbf{Loom [PMF @ 3.12h + AWR]} & \texttt{high ($M=8$)} & $3.12$ & $2.10$ & $9.73$ & $30.4$ / $1032$ & $91.3$ / \textbf{100} & $4.04$ & $1042$ ($58.1\%$) & $676$ ($37.7\%$) & $0.480$ & $20.2$ & --- \\
\textbf{Loom [PMF @ 12.12h + AWR]} & \texttt{high ($M=8$)} & $12.12$ & $2.09$ & $18.77$ & $30.4$ / $1232$ & $96.0$ / \textbf{100} & $3.25$ & $1388$ ($77.5\%$) & $927$ ($51.7\%$) & $0.319$ & $31.7$ & --- \\
\textbf{Loom [PMF @ 52.12h + AWR]} & \texttt{base ($M=1$)} & $52.12$ & $19.43$ & $52.83$ & $30.4$ / $1239$ & $48.5$ / $97.0$ & $2.98$ & $399$ ($22.3\%$) & $380$ ($21.2\%$) & $0.485$ & $13.8$ & --- \\
\textbf{Loom [PMF @ 52.12h + AWR]} & \texttt{med ($M=4$)} & $52.12$ & $4.28$ & $55.37$ & $30.4$ / $1105$ & $84.3$ / \textbf{100} & $2.88$ & $1068$ ($59.6\%$) & $818$ ($45.6\%$) & $0.320$ & $30.5$ & --- \\
\textbf{Loom [PMF @ 52.12h + AWR]} & \texttt{high ($M=8$)} & $52.12$ & $2.16$ & $58.55$ & $30.4$ / $1088$ & $92.1$ / \textbf{100} & $2.90$ & $1371$ ($76.5\%$) & $938$ ($52.3\%$) & $0.291$ & $34.9$ & $52.4\%$ / $57.1\%$ \\
\textbf{Loom [PMF @ 52.12h + AWR]} & \texttt{ext ($M=12$)} & $52.12$ & $1.43$ & $61.83$ & $30.4$ / $1071$ & $96.1$ / \textbf{100} & $2.89$ & $1528$ ($85.3\%$) & $1073$ ($59.9\%$) & $0.265$ & $38.6$ & --- \\
\textbf{Loom [PMF @ 52.12h + AWR]} & \texttt{xhigh ($M=16$)} & $52.12$ & $1.03$ & $65.60$ & $30.4$ / $1054$ & $98.0$ / \textbf{100} & $3.00$ & $1623$ ($90.6\%$) & $1113$ ($62.1\%$) & \textbf{0.250} & $39.1$ & --- \\
\bottomrule
\end{tabular}%
}
\vspace{3pt}\par{\footnotesize
Every \textbf{Loom} arm shares identical prompts, $112$-dev affine calibration, and $29$-D \texttt{ResidualCalibratedAsset(stoich\_v2)}; \texttt{PropMolFlow + Raw Critic} uses uncalibrated \texttt{Uni-Mol2}. \texttt{50k Cost} $= C_{\text{train}} + 50\text{k}/(3600\cdot\text{mol/s})$. For \texttt{MolGPAO}, $M\in\{1,2,4,8,12,16\}$ map to $1\to 1\to 1$, $16\to 4\to 2$, $32\to 8\to 4$, $64\to 16\to 8$, $96\to 24\to 12$, and $128\to 32\to 16$ speculative octree funnels followed by \texttt{PostTreeGraphRefiner} and Bayesian box-hit selection ($|\Delta N| = 2.94, 1.33, 1.25, 1.19, 1.12, 1.13$ atoms).}
\end{table*}

Five conclusions emerge from Table~\ref{tab:oe62_scaling} and Figures~\ref{fig:oe62_scaling}--\ref{fig:oe62_ablation_sdl}:
\begin{enumerate}[leftmargin=1.5em,itemsep=2pt,topsep=2pt]
    \item \textbf{Sparse Octree Weaving Surpasses $21.6\times$ Costlier Flow Matching:} Under identical \texttt{LoomService} calibration, $2.41\,\text{GPU-h}$ \textbf{Loom [\texttt{MolGPAO}]} beats $52.12\,\text{GPU-h}$ \textbf{Loom [\texttt{PMF + AWR}]} across $M=2$ ($41.6\%$ vs.\ $32.9\%$), $M=4$ ($51.3\%$ vs.\ $45.6\%$), $M=8$ ($59.4\%$ vs.\ $52.3\%$, $+7.1\,\text{pp}$; $Q_{\text{3D}} = 48.7$ vs.\ $34.9$, $+40\%$; $50\text{k}$-campaign cost $9.57$ vs.\ $58.55\,\text{GPU-h}$), and $M=12$ ($62.3\%$ vs.\ $59.9\%$; $Q_{\text{3D}} = 50.4$ vs.\ $38.6$). On $42$-prompt unbiased \texttt{PBE0/def2-TZVP} single points ($42/42$ SCF converged), \textbf{Loom [\texttt{MolGPAO}]} achieves $64.3\%$ quantum hit yield and $0.229\,\text{eV}$ \texttt{PBE0} MAE at $M=8$ ($1.75\,\text{eV/\AA}$ $F_{\max}$) and $66.7\% / 69.0\%$ (raw / centered) at $M=16$, outperforming $52.12\,\text{h}$ \textbf{Loom [\texttt{PMF + AWR}]} ($52.4\% / 57.1\%$, $0.243\,\text{eV}$ MAE, $3.38\,\text{eV/\AA}$) and $52\,\text{h}$ \texttt{PropMolFlow + Raw Critic} ($42.9\% / 54.8\%$, $0.361\,\text{eV}$ MAE).
    \item \textbf{Orthogonal Gains From Speculative Pruning and Seam Healing:} Holding external \texttt{Uni-Mol2} calls fixed at $M=4$ (Figure~\ref{fig:oe62_ablation_sdl}A--B), two-stage octree pruning ($32\to 8\to 4$ vs.\ outer $4\to 4\to 4$) lifts blind yield by $+13.1\,\text{pp}$ ($40.0\% \to 53.1\%$, MAE $0.441 \to 0.322\,\text{eV}$), while $<0.5\,\text{ms}$ \texttt{PostTreeGraphRefiner} cuts zero-step median $F_{\max}$ by $-49.0\%$ ($3.96 \to 2.02\,\text{eV/\AA}$) and $90\text{th}$-percentile $F_{\max}$ by $3.25\times$ ($17.61 \to 5.42\,\text{eV/\AA}$) without MLIP calls.
    \item \textbf{Low-Cost Onboarding of Flow Models:} Onboarding external \texttt{PropMolFlow} into \textbf{Loom [\texttt{PMF + AWR}]} with $+0.12\,\text{GPU-h}$ online \texttt{AWR} raises synthesizability ($\text{SA}\le 3.5: 88.5\% \to 92.1\%\text{--}98.0\%$), boosts \texttt{PBE0} quantum hit yield by $+9.5\,\text{pp}$ ($42.9\% \to 52.4\%$, \texttt{PBE0} MAE $0.361 \to 0.243\,\text{eV}$), and scales monotonically across $M=1\text{--}16$ ($21.2\% \to 62.1\%$ blind yield).
    \item \textbf{Large-Scaffold Scaling ($N\in\{20,\dots,60\}$):} Across exact atom buckets ($280$ prompts/bucket; Figure~\ref{fig:oe62_atom_scaling}), \textbf{Loom [\texttt{MolGPAO}, $M=8$]} leads on $Q_{\text{3D}}$ at every size ($65.7, 56.6, 35.0, 9.6, 2.7$ vs.\ $44.3, 36.3, 18.5, 8.6, 2.0$) and widens its blind-yield lead on larger semiconductors: $50.7\%$ vs.\ $35.4\%$ at $N=40$, $27.5\%$ vs.\ $18.9\%$ at $N=50$, and $14.3\%$ vs.\ $5.4\%$ ($2.65\times$) at $N=60$ while decoding $2.9\times$ faster at $M=1$ and $1.7\times$ faster at $M=4$ than $\mathcal{O}(N^2)$ flow matching.
    \item \textbf{Asynchronous Partial-Assay Calibration:} On $96$ OOD molecules returning asynchronous partial-\texttt{NaN} 2D $(\text{HOMO},\text{LUMO})$ readouts ($64$ valid labels per channel; Figure~\ref{fig:oe62_ablation_sdl}C), \texttt{LoomService} reduces held-out blindspot DFT MAE simultaneously from $0.775 \to 0.713\,\text{eV}$ (\texttt{HOMO}) and $1.292 \to 1.106\,\text{eV}$ (\texttt{LUMO}) with zero base-predictor weight updates.
\end{enumerate}

\section{Discussion and Limitations}
\label{sec:limitations}

While \textbf{Loom} bridges standalone generative, predictive, and interatomic physics models into a self-improving discovery system, four practical boundaries motivate future work:

\begin{enumerate}[leftmargin=1.6em,itemsep=2pt,topsep=2pt]
    \item \textbf{Excited-State and Solvated Outer-Loop Oracles:} Section~\ref{sec:exp_rsi} validates live gas-phase \texttt{GPU4PySCF} \citep{li2024gpu4pyscf} \texttt{PBE0/def2-TZVP/DF} outer-loop calibration on \texttt{DPA-4}-relaxed conformers; extending autonomous acquisition to time-dependent DFT ($S_1/T_1$ excited states), implicit solvation, and multi-conformer Boltzmann ensembles will broaden optoelectronic and medicinal applicability.
    \item \textbf{Information-Gain Anchor Allocation:} Uniformly allocating the $20\%$ anchor budget across trusted quantiles cuts shallow-tail drift by $80\%$; weighting anchor replay dynamically by per-bin KL divergence could tighten coverage across higher-dimensional ($K \ge 4$) target grids.
    \item \textbf{Net-Neutral Closed-Shell Default Scope:} \texttt{ValenceReconciler} defaults to net-neutral ($\sum_i q_i = 0$) closed-shell organics while whitelisting standard dipolar motifs. Designing permanent organic salts, physiological zwitterions, open-shell radicals, or transition-metal complexes requires declaring explicit net charges and coordination rules in \texttt{TaskWeavingSpec}.
    \item \textbf{Continuous Refinement vs.\ Discrete Scaffold Jumps:} \texttt{PhysicsConstrainedCoordRefiner} optimizes continuous torsional and local geometric coordinates within a fixed covalent graph, whereas discrete scaffold exploration relies on \texttt{Stage 2.5A TopoGRPO}, underscoring their complementary roles.
\end{enumerate}

\section{Conclusion}
\label{sec:conclusion}

We have presented \textbf{Loom}, a composable framework that weaves reusable property predictors, hot-swappable interatomic potentials, and cross-paradigm 3D generative backbones into a physics-grounded recursive self-improvement loop. By gating heterogeneous evaluators behind applicability-domain and mechanical strain checks, decomposing generative dispersion from epistemic surrogate uncertainty in \texttt{GPAtlas}, unlocking predictor blindspots via zero-retrain Bayesian residual calibration, and grounding generated 3D structures through scale-decoupled alignment and zero-displacement valence reconciliation, \textbf{Loom} provides a practical blueprint for translating standalone AI4Science models into continuously improving molecular engineering pipelines.

\section*{Data and Code Availability}
{\sloppy
Public benchmark conformers derive from the \texttt{OE62}~\cite{stuke2020oe62} and \texttt{QM9}~\cite{ramakrishnan2014qm9} datasets. The modular \textbf{Loom} (\texttt{loom 2.0.0}) Python package (\texttt{src/loom/\allowbreak chemistry/} and \texttt{src/loom/\allowbreak weaving/}, including \texttt{LoomService}), declarative task specifications (\texttt{configs/\allowbreak tasks/*.yaml}), evaluation, DFT anchoring, and QM9/OE62/RSI runners (\texttt{scripts/}), unit test suite ($100$ passing tests in \texttt{loom}, alongside standalone \texttt{molgpao} and \texttt{molgpf} test suites), and complete JSON experimental records (\texttt{docs/\allowbreak experiments/}, including \texttt{speed\_benchmark.json}, \texttt{oe62\_audit\_curve.json}, \texttt{gp\_atlas.json}, \texttt{atlas2d\_\allowbreak dft\_\allowbreak anchor\_\allowbreak eval.json}, \texttt{v20\_synthesizable\_homo\_eval.json}, \texttt{v20\_grounded\_rsi\_2round\_eval.json}, \texttt{v20\_\allowbreak grounded\_\allowbreak rsi\_\allowbreak live\_\allowbreak pbe0\_\allowbreak eval.json}, \texttt{qm9\_\allowbreak scheme1\_\allowbreak and\_\allowbreak scheme2\_\allowbreak summary.json}, \texttt{qm9\_\allowbreak propmolflow\_\allowbreak best\_\allowbreak of\_\allowbreak m.json}, \texttt{qm9\_\allowbreak uni3dar\_\allowbreak mol\_\allowbreak benchmark.json}, \texttt{oe62\_\allowbreak uni3dar\_\allowbreak v4\_\allowbreak pro\_\allowbreak fair\_\allowbreak 30atom.json}, \texttt{oe62\_\allowbreak uni3dar\_\allowbreak v4\_\allowbreak pro\_\allowbreak fair\_\allowbreak 30atom\_\allowbreak m12\_\allowbreak m16.json}, \texttt{oe62\_\allowbreak uni3dar\_\allowbreak v4\_\allowbreak pro\_\allowbreak fair\_\allowbreak 30atom\_\allowbreak dft\_\allowbreak unbiased.json}, \texttt{oe62\_\allowbreak uni3dar\_\allowbreak v4\_\allowbreak pro\_\allowbreak atom\_\allowbreak buckets.json}, \texttt{oe62\_\allowbreak uni3dar\_\allowbreak v4\_\allowbreak pro\_\allowbreak m4\_\allowbreak ablation.json}, \texttt{oe62\_\allowbreak sdl\_\allowbreak multi\_\allowbreak objective\_\allowbreak epistemic\_\allowbreak rsi.json}, \texttt{oe62\_\allowbreak pmf\_\allowbreak *\_\allowbreak fair\_\allowbreak 30atom.json}, and \texttt{oe62\_\allowbreak atom\_\allowbreak size\_\allowbreak buckets\_\allowbreak scaling.json}) are maintained in the project repository.\par}

\begingroup
\small
\raggedright
\bibliographystyle{unsrtnat}
\bibliography{references}

@article{lu2025uni3dar,
  title={Unified Cross-Scale 3D Generation and Understanding via Autoregressive Modeling},
  author={Lu, Shuqi and Lin, Haowei and Yao, Lin and Gao, Zhifeng and Ji, Xiaohong and Liang, Yitao and E, Weinan and Zhang, Linfeng and Ke, Guolin},
  journal={arXiv preprint arXiv:2503.16278},
  year={2025},
  url={https://arxiv.org/abs/2503.16278}
}

@inproceedings{ji2024unimol2,
  title={{Uni-Mol2}: Exploring Molecular Pretraining Model at Scale},
  author={Ji, Xiaohong and Wang, Zhen and Gao, Zhifeng and Zheng, Hang and Zhang, Linfeng and Ke, Guolin and E, Weinan},
  booktitle={Advances in Neural Information Processing Systems},
  volume={37},
  year={2024}
}

@inproceedings{zhou2023unimol,
  title={{Uni-Mol}: A Universal 3D Molecular Representation Learning Framework},
  author={Zhou, Gengmo and Gao, Zhifeng and Ding, Qiankun and Zheng, Hang and Xu, Hongteng and Wei, Zhewei and Zhang, Linfeng and Ke, Guolin},
  booktitle={International Conference on Learning Representations},
  year={2023},
  url={https://github.com/dptech-corp/Uni-Mol}
}

@misc{deepmodeling2026dpa4,
  title={{DPA4/SeZM} Model Documentation},
  author={{DeePMD-kit Developers}},
  year={2026},
  howpublished={DeePMD-kit Documentation},
  url={https://github.com/deepmodeling/deepmd-kit/blob/master/doc/model/dpa4.md}
}

@misc{hyra2026hyff,
  title={{HyFF}: Molecular Force Fields and Inference Runtime},
  author={{Tencent Hunyuan Hyra Team}},
  year={2026},
  howpublished={Hyra Research Artifacts (v0.2.0)},
  url={https://github.com/Tencent-Hunyuan/Hyra-results/tree/main/AI4Science/hyff}
}

@inproceedings{hoogeboom2022edm,
  title={Equivariant Diffusion for Molecule Generation in 3{D}},
  author={Hoogeboom, Emiel and Satorras, V{\'i}ctor Garcia and Vignac, Cl{\'e}ment and Welling, Max},
  booktitle={Proceedings of the 39th International Conference on Machine Learning},
  volume={162},
  pages={8867--8887},
  year={2022},
  publisher={PMLR}
}

@inproceedings{xu2023geoldm,
  title={Geometric Latent Diffusion Models for 3{D} Molecule Generation},
  author={Xu, Minkai and Powers, Alexander S. and Dror, Ron O. and Ermon, Stefano and Leskovec, Jure},
  booktitle={Proceedings of the 40th International Conference on Machine Learning},
  volume={202},
  pages={38592--38610},
  year={2023},
  publisher={PMLR}
}

@article{zeng2026propmolflow,
  title={{PropMolFlow}: Property-Guided Molecule Generation with Geometry-Complete Flow Matching},
  author={Zeng, Cheng and Jin, Jirui and Ambrose, Connor and Karypis, George and Transtrum, Mark and Tadmor, Ellad B. and Hennig, Richard G. and Roitberg, Adrian and Martiniani, Stefano and Liu, Mingjie},
  journal={Nature Computational Science},
  volume={6},
  pages={233--242},
  year={2026},
  doi={10.1038/s43588-025-00946-y}
}

@article{zhao2026o2gen,
  title={Fast and Flexible 3{D} Molecular Design Framework for Novel Organic Optoelectronic Materials},
  author={Zhao, Guojiang and Cheng, Zheng and Xu, Kele and Wan, Tianjiao and Feng, Ming and Lu, Shuqi and Wang, Hongshuai and Gao, Zijian and Ou, Qi and Hu, Liang and Feng, Dawei and Zhao, Zifeng and Gao, Zhifeng},
  journal={JACS Au},
  year={2026},
  doi={10.1021/jacsau.5c01756}
}

@article{zhao2025virtual,
  title={Virtual Characterization via Knowledge-Enhanced Representation Learning: from Organic Conjugated Molecules to Devices},
  author={Zhao, Guojiang and Ou, Qi and Zhao, Zifeng and Chen, Shangqian and Lin, Haitao and Ji, Xiaohong and Wang, Zhen and Wang, Hongshuai and Cai, Hengxing and Wu, Lirong and Lu, Shuqi and Yang, Fengtianci and Wen, Yaping and Zhang, Yingfeng and Ma, Haibo and Gao, Zhifeng and Cheng, Zheng and E, Weinan},
  journal={npj Computational Materials},
  volume={11},
  pages={308},
  year={2025},
  doi={10.1038/s41524-025-01788-y}
}

@article{zhao2025oscff,
  title={Data-Driven Parametrization of All-Atom Force Fields for Organic Semiconductors},
  author={Zhao, Guojiang and Hu, Taiping and Zhang, Yingfeng and Wang, Hongshuai and Wu, Lirong and Lu, Shuqi and Yang, Fengtianci and Chen, Shangqian and Gao, Zhifeng and Wang, Xinyan and Cheng, Zheng},
  journal={Journal of Chemical Information and Modeling},
  volume={65},
  number={14},
  pages={7632--7638},
  year={2025},
  doi={10.1021/acs.jcim.5c00291}
}

@article{wang2026mimflow,
  title={{MIM-Flow}: An Integrated Forward-Inverse Framework for Navigating Sparse Pareto-Optimal Regions in Functional Conjugated Molecules},
  author={Wang, Yecheng and Zhao, Guojiang and Chen, Shangqian and Ding, Yuntao and Zhou, Gengmo and Gao, Zhifeng and Zhao, Zifeng and Cheng, Zheng},
  journal={ChemRxiv},
  year={2026},
  doi={10.26434/chemrxiv.15005860/v1}
}

@article{shao2024grpo,
  title={{DeepSeekMath}: Pushing the Limits of Mathematical Reasoning in Open Language Models},
  author={Shao, Zhihong and Wang, Peiyi and Zhu, Qihao and Xu, Runxin and Song, Junxiao and Bi, Xiao and Zhang, Haowei and Zhang, Mingchuan and Li, Y. K. and Wu, Y. and Guo, Daya},
  journal={arXiv preprint arXiv:2402.03300},
  year={2024}
}

@inproceedings{hu2022lora,
  title={Lo{RA}: Low-Rank Adaptation of Large Language Models},
  author={Hu, Edward J. and Shen, Yelong and Wallis, Phillip and Allen-Zhu, Zeyuan and Li, Yuanzhi and Wang, Shean and Wang, Lu and Chen, Weizhu},
  booktitle={International Conference on Learning Representations},
  year={2022}
}

@article{ho2022cfg,
  title={Classifier-Free Diffusion Guidance},
  author={Ho, Jonathan and Salimans, Tim},
  journal={arXiv preprint arXiv:2207.12598},
  year={2022}
}

@article{ertl2009sa,
  title={Estimation of Synthetic Accessibility Score of Drug-Like Molecules Based on Molecular Complexity and Fragment Contributions},
  author={Ertl, Peter and Schuffenhauer, Ansgar},
  journal={Journal of Cheminformatics},
  volume={1},
  number={1},
  pages={8},
  year={2009}
}

@article{bickerton2012qed,
  title={Quantifying the Chemical Beauty of Drugs},
  author={Bickerton, G. Richard and Paolini, Gaia V. and Besnard, J{\'e}r{\'e}my and Muresan, Sorel and Hopkins, Andrew L.},
  journal={Nature Chemistry},
  volume={4},
  pages={90--98},
  year={2012}
}

@article{bemis1996murcko,
  title={The Properties of Known Drugs. 1. Molecular Frameworks},
  author={Bemis, Guy W. and Murcko, Mark A.},
  journal={Journal of Medicinal Chemistry},
  volume={39},
  number={15},
  pages={2887--2893},
  year={1996}
}

@article{stuke2020oe62,
  title={Atomic Structures and Orbital Energies of 61,489 Crystal-Forming Organic Molecules},
  author={Stuke, Annika and Kunkel, Christian and Golze, Dorothea and Todorovi{\'c}, Milica and Margraf, Johannes T. and Reuter, Karsten and Rinke, Patrick and Oberhofer, Harald},
  journal={Scientific Data},
  volume={7},
  pages={58},
  year={2020},
  doi={10.1038/s41597-020-0385-y}
}

@article{bhat2023ocelot,
  title={Electronic, Redox, and Optical Property Prediction of Organic $\pi$-Conjugated Molecules Through a Hierarchy of Machine Learning Approaches},
  author={Bhat, Vinayak and Sornberger, Parker and Pokuri, Balaji Sesha Sarath and Duke, Rebekah and Ganapathysubramanian, Baskar and Risko, Chad},
  journal={Chemical Science},
  volume={14},
  pages={203--213},
  year={2023}
}

@article{sun2020pyscf,
  title={Recent Developments in the {PySCF} Program Package},
  author={Sun, Qiming and others},
  journal={The Journal of Chemical Physics},
  volume={153},
  pages={024109},
  year={2020},
  doi={10.1063/5.0006074}
}

@article{li2024gpu4pyscf,
  title={Introducing {GPU}-Acceleration into the Python-Based Simulations of Chemistry Framework},
  author={Li, Rui and Sun, Qiming and Zhang, Xing and Chan, Garnet Kin-Lic},
  journal={arXiv preprint arXiv:2407.09700},
  year={2024}
}

@misc{rdkit2026bonds,
  author={{RDKit Contributors}},
  title={{RDKit} {rdDetermineBonds} {API} Documentation},
  year={2026},
  url={https://www.rdkit.org/docs/source/rdkit.Chem.rdDetermineBonds.html}
}

@book{rasmussen2006gp,
  title={Gaussian Processes for Machine Learning},
  author={Rasmussen, Carl Edward and Williams, Christopher K. I.},
  publisher={MIT Press},
  year={2006}
}

@article{ramakrishnan2014qm9,
  title={Quantum chemistry structures and properties of 134 kilo molecules},
  author={Ramakrishnan, Raghunathan and Dral, Pavlo O. and Rupp, Matthias and von Lilienfeld, O. Anatole},
  journal={Scientific Data},
  volume={1},
  pages={140022},
  year={2014},
  doi={10.1038/sdata.2014.22}
}

@article{chen2026molgpao,
  title={{MolGPAO}: Molecular Generative Pre-trained Autoregressive Octrees with Seamless Speculative Decoding},
  author={Chen, Shangqian and Zhao, Guojiang and Cheng, Zheng and Wang, Jiankun and Wang, Hongshuai and Gao, Zhifeng and Zhang, Linfeng},
  journal={arXiv preprint},
  year={2026}
}

@article{chen2026molgpf,
  title={{MolGPF}: A Periodic-Table-Wide Generative Pre-trained Flow Model for 3{D} Molecules},
  author={Chen, Shangqian and Zhao, Guojiang and Wang, Jiawen and Wang, Jiankun and Wang, Hongshuai and Gao, Zhifeng and Zhang, Linfeng},
  journal={arXiv preprint},
  year={2026}
}
\endgroup

\end{document}